\documentclass[lettersize,journal]{IEEEtran}
\usepackage{amsmath,amsfonts}
\usepackage{algorithmic}
\usepackage{algorithm}
\usepackage{array}
\usepackage{subfigure}
\usepackage[caption=false,font=normalsize,labelfont=sf,textfont=sf]{subfig}

\usepackage{url}
\usepackage{graphicx}
\usepackage{booktabs}
\usepackage{multirow}
\usepackage{dsfont}

\begin{document}
\title{Learning Discriminative Representations and Decision Boundaries for Open Intent Detection}
\author{Hanlei Zhang, Hua Xu, Shaojie Zhao, Qianrui Zhou
	\thanks{Hanlei Zhang, Hua Xu, Shaojie Zhao and Qianrui Zhou are with the State Key Laboratory of Intelligent Technology and Systems, Department of Computer Science and Technology, 
		Tsinghua University, Beijing 100084, China (e-mail:zhang-hl20@mails.tsinghua.edu.cn; xuhua@tsinghua.edu.cn; murrayzhao@163.com;  zgr22@mails.tsinghua.edu.cn).
	}
	\thanks{Shaojie Zhao is with the State Key Laboratory of Intelligent Technology and Systems, Department of Computer Science and Technology, Tsinghua University, Beijing 100084, China, and also with the School of Information Science and Engineering, Hebei University of Science and Technology, Shijiazhuang 050018, China (e-mail: murrayzhao@163.com).
	}
        \thanks{
        The full data and codes are available for use at https://github.com/thuiar/TEXTOIR.
        }
}	

\maketitle

\begin{abstract}
Open intent detection is a significant problem in natural language understanding, which aims to identify the unseen open intent while ensuring known intent identification performance. However, current methods face two major challenges. Firstly, they struggle to learn friendly representations to detect the open intent with prior knowledge of only known intents. Secondly, there is a lack of an effective approach to obtaining specific and compact decision boundaries for known intents. To address these issues, this paper presents an original framework called DA-ADB, which successively learns distance-aware intent representations and adaptive decision boundaries for open intent detection. Specifically, we first leverage distance information to enhance the distinguishing capability of the intent representations. Then, we design a novel loss function to obtain appropriate decision boundaries by balancing both empirical and open space risks. Extensive experiments demonstrate the effectiveness of the proposed distance-aware and boundary learning strategies. Compared to state-of-the-art methods, our framework achieves substantial improvements on three benchmark datasets. Furthermore, it yields robust performance with varying proportions of labeled data and known categories.
\end{abstract}


\section{Introduction}
\IEEEPARstart{I}{ntent} detection plays a critical role in natural language understanding (NLU), aiming to mine user purposes behind the text utterances. The traditional intent detection task is restricted to closed-world classification. It assumes all the intent categories are accessible and has achieved great progress with a booming of effective methods for supervised classification~\cite{ijcai2021-0622,weld2022survey}.

\begin{figure}[t!]
	\centering  
	\includegraphics[width=0.95\columnwidth]{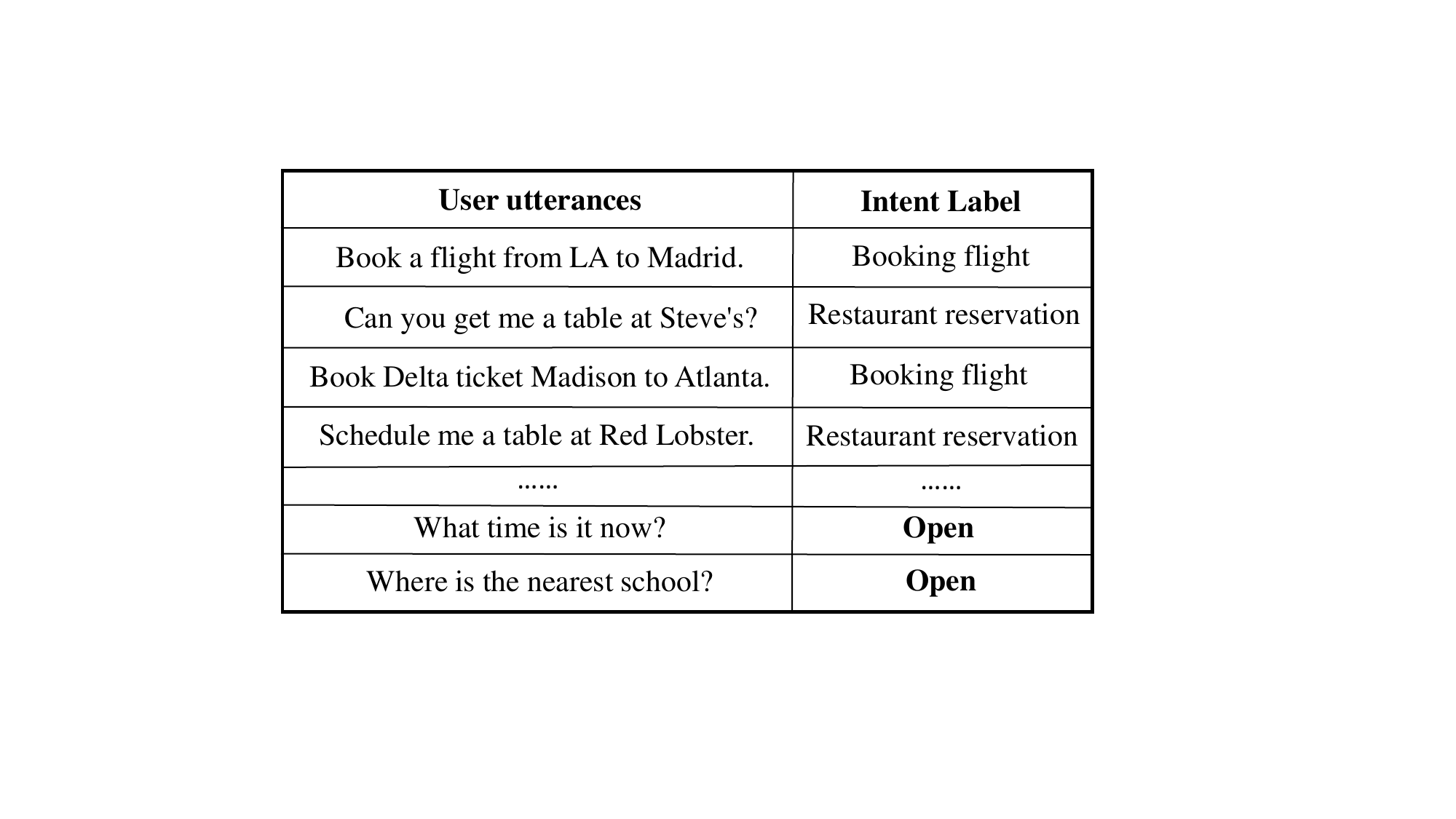}
	\caption{\label{example} An example of open intent detection. Booking flight and Restaurant reservation are two known intents. We should identify them correctly while detecting the utterances with the open intent.}
\end{figure}

Nonetheless, due to the variety and uncertainty of the user needs, it is usually inapplicable to cover all intent categories. Taking Fig~\ref{example} as an example, there are two task-specific intents of booking flight and restaurant reservation. Ideally, we hope to identify each utterance within the two known intent categories. However, in the real application, some unexpected utterances may exist with unknown intents that have never been seen before, such as asking about the time or place. Effectively detecting these unknown intents helps reduce false-positive errors and makes the system more robust. Moreover, we can leverage them to explore more potential user needs and improve customer satisfaction. 

Our previous works first proposed the open (unknown) intent detection task~\cite{lin-xu-2019-deep,lin2019post,Zhang_Xu_Lin_2021} to solve this problem. As we do not have any prior knowledge of unknown intents (e.g.,  the specific categories and the class number), all the unknown classes are regarded as one open class. Specifically, the goal of this task is to use the prior knowledge of only $K$-class known intents to detect the unseen ($K$+1)$^{\text{th}}$ class open intent while ensuring the known intent identification performance. It requires no need to collect labeled data for each fine-grained open intent category during training or evaluation, saving much time and workforce for practical application. 

Similar open set problems were first explored in computer vision~\cite{9040673}. Among these problems, open set recognition (OSR)~\cite{scheirer2013toward} has the closest setting to our task, which also aims to identify known classes and reject the unknown class that does not appear in the training set. However, the setting of OSR can use unknown-class data for tuning parameters~\cite{9040673}. In contrast, the open intent data are unavailable during validation in our task, which is more suitable in real-world scenarios. Besides, OSR methods are only applied in visual tasks, which may not work well on discrete text data~\cite{bendale2016towards,9521769}. Fei and Liu~\cite{fei-liu-2016-breaking} extended this problem to the text classification but also used unknown-class data for obtaining SVM boundaries. Shu et al.~\cite{Shu2017DOCDO} adopted the convolution neural network (CNN) to extract deep features and obtained tight confidence thresholds for each known class based on the statistics information. However, separating known classes and the open class with dense confidence scores may be hard. 

Out-of-distribution (OOD) detection is a task that involves identifying outliers that differ from in-distribution (ID) data during testing. The goal is to generate discriminative scores that can distinguish between ID and OOD samples~\cite{hendrycks17baseline,liang2018enhancing,hendrycks2018deep, Kim2018JointLO,xu2021unsupervised,shen2021enhancing}. While our task is similar, the main difference is that we aim to design appropriate decision criteria to balance the performance of multi-class ID and the one-class OOD samples. Our experiments have shown that using OOD detection methods directly after training a model on IND data can lead to a degradation in performance compared to open intent detection methods. 

The research on open intent detection is just beginning in recent years. Lin and Xu~\cite{lin-xu-2019-deep} made the first trial on this task with the feature-based methods. In particular, the large margin cosine loss~\cite{8578650} was first adopted to learn representations with intra-class compactness and inter-class separation properties. Then, the local outlier factor (LOF)~\cite{breunig2000lof} was used to detect the low-density examples as the open class. Nevertheless, the embeddings optimized in cosine space may be less suitable for LOF~\cite{yan-etal-2020-unknown}. Yan et al.~\cite{yan-etal-2020-unknown} leveraged the Gaussian mixture model incorporating the class label information to obtain more suitable representations for anomaly detection. However, the performance drops dramatically when the intent categories have complicated semantics. Recent works tried to construct pseudo open class samples for ($K$+1)-way training ~\cite{zhan-etal-2021-scope,9693239}, yet the samples from the open class may not follow the same distribution in the semantic space.

There are two main difficulties in current open intent detection methods. Firstly, the representations trained on known intents may not be robust enough for detecting the unseen open class. Secondly,  the decision conditions (e.g., confidence or density thresholds) need to be selected manually and are implicit based on the prior knowledge of only known intents.

We propose a novel framework, DA-ADB, for open intent detection to solve these problems, which learns discriminative representations and decision boundaries with only known intents. It first extracts deep intent representations from the pre-trained language model BERT~\cite{devlin2019bert} at the sentence level and calculates the centroids by averaging the samples of each known class. 
Then, it aims to perceive the distance information to learn representations with distinguishing capability. Specifically, to compute the distance-aware coefficient, each sample compares the Euclidean distances between its nearest and next-nearest centroids. The coefficient is incorporated into the original intent representation to produce the meta-embedding. The magnitude of its meta-embedding reflects the hardness of each sample. That is, the larger magnitude corresponds to the easier sample with more confidence to differentiate the two nearest centroids. To achieve this goal, a cosine classifier~\cite{gidaris2018dynamic} is adopted after the meta-embedding to consider the effect of the magnitude information, which helps focus on harder samples with smaller magnitudes during training. In this way, the intent representations are calibrated to distance-aware concepts for more robust performance. 

After representation learning, we aim to obtain specific and compact decision boundaries in the intent feature space. We suppose each known intent cluster is constrained in a spherical decision boundary to its centroid, which helps reduce the open space risk~\cite{fei-liu-2016-breaking}. The decision boundaries are determined by the radius of each ball area and should be flexibly adaptive to different feature distributions. In particular, the boundary parameters are first initialized with standard normal distribution and then projected with a learnable activation function to get the radius of each decision boundary. 

The key factor is how to control the radius to learn compact decision boundaries for open intent detection, which should satisfy two conditions. On the one hand, they should be broad enough to surround known intent samples as much as possible. On the other hand, they need to be tight enough to prevent the open intent samples from being identified as known intents. A new loss function is designed to address these issues, optimizing the boundary parameters by balancing both the open space and empirical risk with known intent samples inside and outside the decision boundaries. With the boundary loss, the decision boundaries can automatically adapt to the intent feature space until balance. The boundary learning process is a post-processing method that requires no modifying the original model architecture. It works even with the features trained on the simple softmax loss. The distance-aware strategy can further facilitate learning more discriminative representations for better performance.

Our contributions are summarized as follows:
\begin{itemize}
	\item We clarify the definition of the open intent detection problem and propose a novel and effective framework DA-ADB for the main challenges.  
	\item A distance-aware strategy is designed to capture the distinguishing ability of each sample, which helps learn discriminative intent features. 
	\item A novel post-processing method is proposed to learn tight decision boundaries adaptive to the feature space. To the best of our knowledge, it is the first attempt to automatically learn adaptive decision boundaries for detecting the unseen open class.
	\item Extensive experiments conducted on three challenging datasets show that our approach achieves consistently better and more robust results than state-of-the-art methods.
\end{itemize}

The idea of adaptive decision boundary (ADB) was presented in a preliminary version of this paper published in the proceeding of the thirty-fifth AAAI conference (AAAI-21)~\cite{Zhang_Xu_Lin_2021}. In this paper, we extend the preliminary version in the following aspects:
\begin{itemize}
	\item A novel method is introduced to incorporate the distance information into the intent representations. It can capture the hardness of each sample with the distance-aware coefficient for effective training. 
	\item A series of experiments are conducted to show the advantages of injecting distance-aware concepts into intent representations for open intent detection.
	\item We formulate the open intent detection problem and enrich the introduction and related work. More baselines are reproduced and added to our experiments with detailed analysis. 
	\item The experimental results are updated with our TEXTOIR platform~\cite{zhang-etal-2021-textoir}, which has standard and unified interfaces for a fair comparison.  
\end{itemize}
\section{Related Works}
This section reviews the related works in open set recognition, out-of-distribution, and open intent detection.
\subsection{Open Set Recognition}
OSR is a pioneering work related to us, aiming to reject the negative samples while identifying positive samples. At first, researchers used SVM-based methods as the solutions. One-class SVM~\cite{scholkopf2001estimating} was designed for binary classification, which found the plane based on the positive training data and regarded the origin as the only member of the negative class. One-vs-all SVM~\cite{Rifkin2004In} was designed for multi-class open classification, which trained the binary classifier for each class and treated the negative classified samples as the open class. Scheirer et al.~\cite {scheirer2013toward} extended the method to computer vision and introduced the concept of open space risk. They proposed the one-vs-set machine to improve generalization ability by compressing the decision space of one-class SVM. Jain et al.~\cite {10.1007/978-3-319-10578-9_26} used a Weibull-calibrated multi-class SVM to estimate the posterior probability satisfying the statistical Extreme Value Theory (EVT). Scheirer et al.~\cite {6809169} presented a Compact Abating Probability (CAP) model, which further improved the performance of Weibull-calibrated SVM by truncating the abating probability.


However, the SVM-based methods have difficulties in capturing advanced pattern semantic concepts~\cite{lin2019post}. Thus, researchers used deep neural networks for OSR. For example,  Bendale and Boult~\cite{bendale2016towards} designed an OpenMax
layer after the penultimate layer of deep neural networks (DNNs) to estimate the open class probability. Zhou et al.~\cite{zhou2021learning} calibrated the closed-set classifier by learning the classifier and data placeholders, which are used to distinguish between known and unknown data and simulate open-class data, respectively. Chen et al.~\cite{9521769} constructed reciprocal points to reduce the empirical risk and further introduced a bounded adversarial mechanism to reduce the open space risk. Nevertheless, these methods only verified the effectiveness of benchmarks in computer vision. Shu et al.~\cite{Shu2017DOCDO} explored this task in natural language processing. They used the output layer of sigmoids and calculated the confidence thresholds based on Gaussian statistics, but the method performs worse when the output probabilities are not discriminative.

\subsection{Out-of-distribution Detection}
OOD detection is a popular task that has received much attention in recent years, which goal is to detect the samples in the testing set that exhibit distribution shifts~\cite{yang2021generalized}. 
A line of works used both ID and ground-truth/generated OOD samples during training. For instance, Kim and Kim~\cite{Kim2018JointLO} jointly trained an in-domain classifier and an out-of-domain detector with both ID and OOD annotated utterances. Hendrycks et al.~\cite{hendrycks2018deep} proposed the use of outlier exposure (OE) to train an OOD detector with external OOD samples. However, annotating OOD samples can be time-consuming and labor-intensive. To address this issue, researchers used the generative adversarial network (GAN) to generate OOD samples. Yu et al.~\cite{Yu2017OpenCategoryCB} adopted adversarial learning to generate positive and negative samples for training the classifier. Ryu et al.~\cite{ryu-etal-2018-domain} used a GAN to train on the ID samples and detected OOD samples with a discriminator. Zheng et al.~\cite{9052492} used a GAN-based generator to produce pseudo OOD samples of discrete token sequences. Nevertheless, it has been shown that deep generative models have limitations in learning high-level semantics on discrete text data~\cite{2018arXiv181009136N}.

Another line of works used only ID samples during training. For example, Hendrycks et al.~\cite{hendrycks17baseline} calculated the softmax probability from ID samples and rejected the low-confidence OOD samples with a threshold. Liang et al.~\cite{liang2018enhancing} used temperature scaling and input pre-processing to enlarge the differences between ID and OOD samples. Xu et al.~\cite{xu2021unsupervised} first fine-tuned the pre-trained transformer with the objective of masked language modeling (MLM) and then utilized the distance information of features from all layers for detecting outliers. Moreover, a series of methods have been developed to design the score function without modifying the network architecture~\cite{lee2018simple,liu2020energy}. However, all of the above methods mainly focus on binary classification and may suffer a  performance decrease when adapting them to our $K$+1 classification task.

\subsection{Open Intent Detection} 
Intent detection is a fundamental task in NLU. For example, the joint slot filling and intent detection task~\cite{zhang-etal-2019-joint,e-etal-2019-novel,qin-etal-2019-stack} has been extensively studied and achieved outstanding performance on standard benchmark datasets~\cite{hemphill-etal-1990-atis,coucke2018snips}. Although more challenging intent benchmark datasets have been proposed in recent years~\cite{larson-etal-2019-evaluation,Casanueva2020,zhang2022mintrec}, they perform intent detection under the assumption of closed world classification~\cite{Casanueva2020}. 

However, in real-world scenarios, user needs are diverse and unpredictable, and it is almost impossible to know all intent categories in advance. Therefore, several approaches have been proposed to detect the open (unknown) intent. For example, Brychcin and Kr{'a}l~\cite{Brychcin2017UnsupervisedDA} proposed an unsupervised method for intent modeling, but it failed to leverage the information of known intents. Zhang et al.~\cite{zhang-etal-2020-discriminative} explored intent detection in few-shot scenarios by learning relations between synthesized positive and negative samples. However, this method needs to augment a large amount of data pairs, which is impractical with more labeled data and leads to unacceptable training and inference time. Xia et al.~\cite{xia-etal-2018-zero} performed intent detection under the zero-shot setting, which assumes the class number and side information of open intents are known. However, there is usually no prior knowledge of open intents in real applications. To address these challenges, we proposed the open intent detection task~\cite{lin-xu-2019-deep,lin2019post,Zhang_Xu_Lin_2021}. In this task, only known intents are used for training and validation. The goal is to identify known intents and detect the one-class open intent during testing. 

Lin and Xu~\cite{lin-xu-2019-deep}  first tackled this problem by learning deep intent features with the margin loss and detected the open intent with LOF. Yan et al.~\cite{yan-etal-2020-unknown} replaced the margin loss with the Gaussian mixture loss to learn better embeddings, but the performance largely depends on the class-label semantics. Moreover, the density-based algorithm is unable to construct specific decision boundaries. Zhan et al.~\cite{zhan-etal-2021-scope} and Cheng et al.~\cite{9693239} regarded different (pseudo) open intents as one class during training, but it may lead to the collapse with intents of disparate semantics.  

\section{Problem Formulation}
\label{problem_formulate}
In open intent detection, we are given an intent label set $I$ and a data set $D$. The intent label set $I=\{I^{\text{Known}}, \textit{Open}\}$, where $I^{\text{Known}} = \{I_{1}, \dots, I_{K}\}$ is the known intent label set and $K$ is the number of known intents. Notably, there may be multiple remaining intent labels in the initial label set  $I \backslash \{I^{\text{Known}}\}$, as indicated in~\cite{lin2020discovering,Zhang_Xu_Lin_Lyu_2021}. In this problem, the non-known intents are all assigned the unified \textit{Open} label. 

The data set $D=\{D^{\text{Train}}, D^{\text{Valid}}, D^{\text{Test}}\}$ consists of training, validation and testing sets. Each subset (e.g., $D^{\text{Train}}$) contains a set of labeled samples $(\boldsymbol{s}_{i}, y_{i})$, where $\boldsymbol{s}_{i}$ is the $i^{\text{th}}$ utterance, and $y_{i}$ is its intent label. 

The intent label set for both $D^{\text{Train}}$ and $D^{\text{Valid}}$ is $I^{\text{Known}}$, while for $D^{\text{Test}}$ is $I$. The training and validation sets contain merely known intent samples, and the unseen open intent samples only exist in the testing set. The goal of open intent detection is to leverage the $K$-class known intents as prior knowledge to both identify known intents and detect the ($K$+1)$^{\text{th}}$ class open intent. 

\begin{figure*}
	\centering
	\includegraphics[scale=.6]{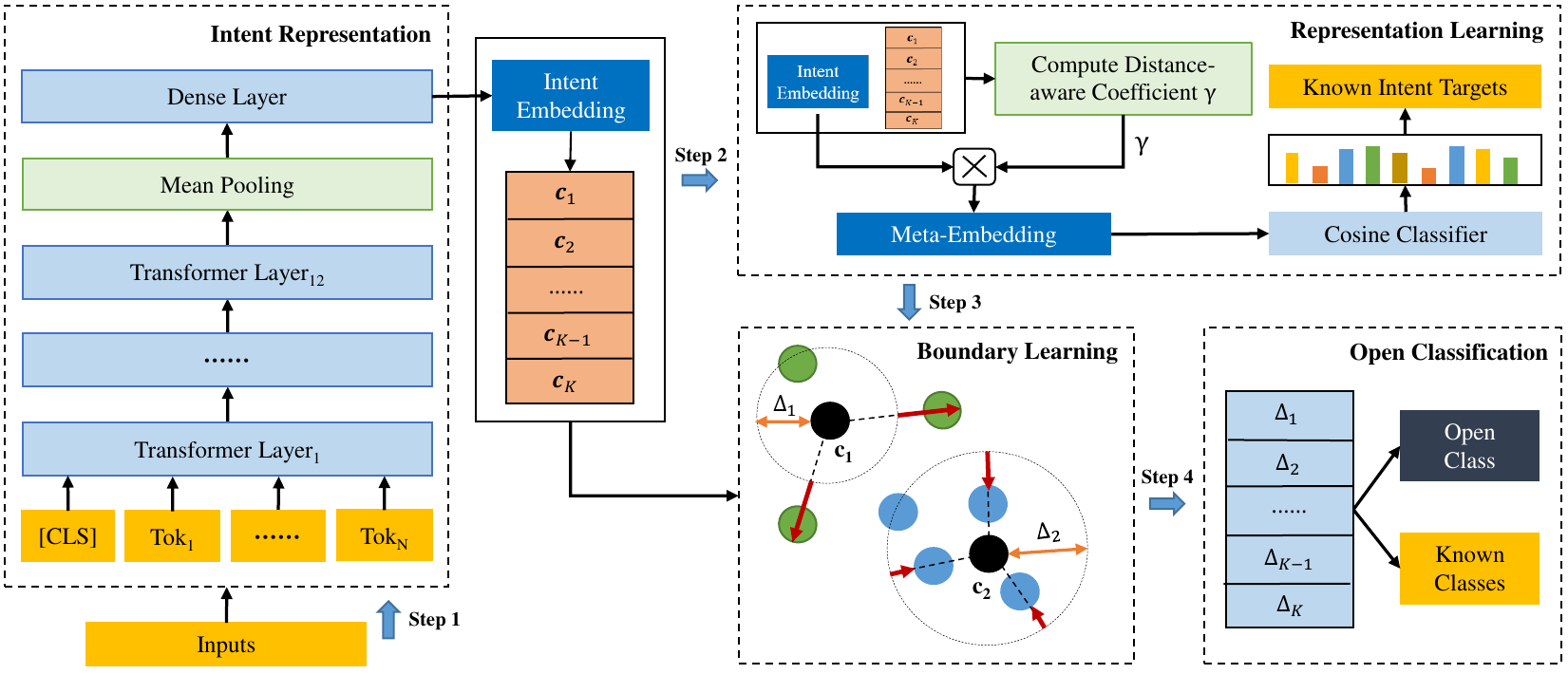}
	\caption{The overall architecture of the proposed framework. Firstly, we use the pre-trained language model BERT to get intent embeddings and average them for each known class to obtain the centroids $\left\{\boldsymbol{c}_{i}\right\}_{i=1}^{K}$. Then, the intent embeddings and centroids are leveraged to compute distance-aware coefficients, further multiplied over the original intent embeddings to yield the meta-embeddings. A cosine classifier learns the distance information with the known intent targets. Next, a new loss function is proposed to learn the radii of the decision boundaries $\left\{\Delta_{i}\right\}_{i=1}^{K}$ adaptive to the intent feature space. Finally, the centroids and decision boundaries are used for open intent detection. 
	}
	\label{model}
\end{figure*}

\section{The Proposed Approach}
This section presents a novel framework for learning friendly intent representations and appropriate decision boundaries for open intent detection. It contains four main steps, intent representation, distance-aware representation learning, adaptive decision boundary learning, and open classification. Fig~\ref{model} shows the overall architecture of our proposed approach. 

\subsection{Intent Representation}
The pre-trained BERT language model is adopted to extract deep intent features. Given the $i^{\text{th}}$ input utterance $\boldsymbol{s}_{i}$, we get all its token embeddings $[CLS, Tok_{1}, \cdots, Tok_{M}]\in \mathds R^{(M+1) \times H}$ from the last hidden layer of BERT. As suggested in~\cite{lin2020discovering}, we  perform mean-pooling on these token embeddings to synthesize the semantic features in the utterance and get the averaged representation $\boldsymbol{x}_{i} \in \mathds R^{H}$:
\begin{align}
\boldsymbol{x}_{i} = \text{mean-pooling}([CLS, Tok_1, \cdots, Tok_M]),
\end{align}
where $CLS$ is a special classification token, $M$ is the sequence length and $H$ is the hidden layer size 768. To further strengthen feature extraction capability, we feed $\boldsymbol{x}_{i}$ to a dense layer $h$ to get the intent representation $\boldsymbol{z}_i \in \mathds R^{D}$:
\begin{align}
\boldsymbol{z}_i=h(\boldsymbol{x}_i) = \sigma(W_h\boldsymbol{x}_{i}+b_h),
\end{align}
where $D$ is the feature dimension, $\sigma$ is the $\operatorname{ReLU}$ activation function, $W_h \in \mathds R^{H \times D}$ and $b_h \in \mathds R^{D}$ respectively denote the weights and the bias term of layer $h$. 
\subsection{Distance-aware Representation Learning}
A new distance-aware representation learning strategy is introduced to learn discriminative intent features. In this method, the centroids of each known class are first calculated and then used to compute the distance-aware coefficient of each sample to obtain the meta-embedding. After the meta-embedding, a cosine classifier enables each sample to perceive the distance information. 
\subsubsection{Centroids Calculation}
\label{centers_def}
To calculate the centroids of each known class, 
let $S=\{(\boldsymbol{z}_i, y_{i}),\ldots, (\boldsymbol{z}_N, y_{N})\}$ be $N$ known intent labeled examples. $S_{k}$ denotes the set of feature vectors labeled with class $k$. The centroid $\boldsymbol{c}_{k} \in \mathds  R^{D}$ is the mean vector of embedded examples in $S_{k}$:
\begin{align}
\boldsymbol{c}_{k}=\frac{1}{\left|S_{k}\right|} \sum_{\left(\boldsymbol{z}_i, y_{i}\right) \in S_{k}} \boldsymbol{z}_i,
\label{center}
\end{align}
where $|S_{k}|$ denotes the number of examples in $S_{k}$. 
\subsubsection{Meta Embedding with Distance-aware Concept}
The initial intent representations have limitations in identifying whether an example is "easy" or "hard" during training, which is unfavorable for discriminative representation learning. To address this issue, we leverage the distance-aware concept to obtain the meta-embedding for enhancing the distinguishing ability. 

For each example, the confidence of which known class it belongs depends on the Euclidean distances between it and known-class centroids in the feature space. The index of the nearest centroid $k_{a}$ is most likely to be its corresponding class, while the index of the next-nearest centroid $k_{b}$ is the most confusing category to be classified. Thus, $k_{a}$ and  $k_{b}$ are the two most informative centroid indexes to evaluate the distinguishing ability, and they are computed by:
\begin{align}
k_{a}=\underset{k}{\operatorname{argmin}}\left\{\|\boldsymbol{z}_{i}-\boldsymbol{c}_{k}\|_{2}\right\}_{k \in I^{\text{Known}}},\\
k_{b}=\underset{k}{\operatorname{argmin}}\left\{\|\boldsymbol{z}_i-\boldsymbol{c}_{k}\|_{2}\right\}_{k \in I^{\text{Known}} \backslash \{k_{a}\},
}
\end{align}
where $\|\boldsymbol{z}_{i}-\boldsymbol{c}_{k}\|_{2}$ denotes the Euclidean distance between $\boldsymbol{z}_{i}$ and $\boldsymbol{c}_k$. A discriminative example should be close to its nearest centroid and far away from the next nearest centroid. Thus, the difference between  $\|\boldsymbol{z}_i-\boldsymbol{c}_{k_{b}}\|_{2}$ and $\|\boldsymbol{z}_i-\boldsymbol{c}_{k_{a}}\|_{2}$ is used to reflect the separating capacity of $\boldsymbol{z}_{i}$. In particular, the distance-aware coefficient $\gamma_{i}$   is defined as:
\begin{align}
\gamma_{i} =  \exp(\|\boldsymbol{z}_i-\boldsymbol{c}_{k_{b}}\|_{2}-\|\boldsymbol{z}_i-\boldsymbol{c}_{k_{a}}\|_{2}) \text{ s.t. } \gamma_{i} \geq 1,
\end{align} 
where $\exp(\cdot)$ enables an exponentially large reception field. It enhances the effect of differentiation and avoids the trivial solution when $\|\boldsymbol{z}_i-\boldsymbol{c}_{k_{b}}\|_{2}$ is close to $\|\boldsymbol{z}_i-\boldsymbol{c}_{k_{a}}\|_{2}$.  Particularly, $\gamma_{i}$ also suggests the difficulty of an example. An "easy" example is more confident to distinguish between the two nearest centroids and has a large $\gamma_{i}$. Yet, a "hard" example is more likely to be confused by the next nearest centroid and has a small $\gamma_{i}$. 

To leverage the distance-aware concept, the intent representation $\boldsymbol{z}_{i}$ is multiplied by  $\gamma_{i}$ to obtain the meta-embedding $\boldsymbol{z}_{i}^{meta}$:
\begin{align}
\boldsymbol{z}_{i}^{meta}=\gamma_{i} \cdot \boldsymbol{z}_{i}.
\end{align}

\subsubsection{Representation Learning}
\label{alpha}
As the distance-aware coefficient is positively correlated with the magnitude of meta-embedding, it is natural to use the vector length to represent the distance-aware concept. For this purpose, the cosine classifier~\cite{gidaris2018dynamic} is adopted to capture the distance information contained in the meta-embedding. 

Specifically, the cosine similarity operator is used on the normalized meta-embeddings and weight vectors to compute the classification logits:
\begin{align}
\phi(\boldsymbol{z}_{i}^{meta})^{k}=\alpha \cdot \cos \left(\boldsymbol{z}_{i}^{meta}, w_{k}^\star\right)=\alpha \cdot \overline{\boldsymbol{z}_{i}^{meta}}^{\top} \overline{w_{k}^\star},
\end{align}
where $\phi(\cdot)$ is the cosine classifier and $\phi(\cdot)^k$ are the output logits of the $k^{\text{th}}$ class, $\alpha$ is a scalar hyper-parameter  (detailed discussion can be seen in section~\ref{analysis_of_alpha}). $\cos(\cdot)$ is the cosine similarity operator, which first normalizes the meta-embedding $\boldsymbol{z}_{i}^{meta}$ and the $k^{\text{th}}$ class weight vector $w_{k}^\star$ and then performs dot product operation. 

In particular,  $\boldsymbol{z}_{i}^{meta}$ and $w_{k}^\star$ are applied by a non-linear squashing function~\cite{sabour2017dynamic} and L2 normalization to obtain $\overline{\boldsymbol{z}_{i}^{meta}}$ and $\overline{w_{k}^\star}$, respectively:
\begin{align}
\overline{\boldsymbol{z}_{i}^{meta}}=\frac{\left\|\boldsymbol{z}_{i}^{meta}\right\|^{2}}{1+\left\|\boldsymbol{z}_{i}^{meta}\right\|^{2}} &\frac{\boldsymbol{z}_{i}^{meta}}{\left\|\boldsymbol{z}_{i}^{meta}\right\|},\\
\overline{w_{k}^\star} = \frac{w_{k}^\star}{||w_{k}^\star||},
\end{align}
where $\|\boldsymbol{z}_{i}^{meta}\|$ and $\|w_{k}^\star\|$ denote the magnitudes of $\boldsymbol{z}_{i}^{meta}$ and $w_{k}^\star$. The non-linear squashing function is helpful to reflect the magnitude information of a vector. It shrinks  $\boldsymbol{z}_{i}^{meta}$ with a large magnitude to the length slightly below 1 and with a short magnitude to the length of almost 0. The L2 normalization eliminates the effect of the weight vector magnitudes on classification logits.

The cosine classifier converts the magnitude information to discriminative classification outputs and facilitates paying more attention to the "hard" example, which outputs lower classification scores with a smaller magnitude.

Finally, we use the softmax loss $	\mathcal{L}_{s}$ to train the meta-embeddings  under the supervision of  known intent targets:
\begin{align}
\mathcal{L}_{s}=-\frac{1}{N}\sum_{i=1}^{N} \log\frac{\exp(\phi( \boldsymbol{z}_{i}^{meta})^{y_{i}})}{\sum_{j=1}^{K}\exp(\phi(\boldsymbol{z}_{i}^{meta})^{j})},
\end{align}
where $y_{i}$ is the label of the $i^{\text{th}}$ example. The initial intent representations can be calibrated to distance-aware concepts during training, and they are further used for learning decision boundaries.

\subsection{Adaptive Decision Boundary Learning}
An original approach to learning the adaptive decision boundary (ADB) is designed for open intent detection. We first formulate the decision boundary and then propose our boundary learning strategy for optimization. 

\subsubsection{Decision Boundary Formulation}
It has been shown the superiority of the spherical shape boundary for open world classification~\cite{fei-liu-2016-breaking}, which greatly reduced the open space risk compared with the half-space binary linear classifier~\cite{scholkopf2001estimating} and two parallel hyper-planes~\cite{scheirer2013toward}. 
Inspired by this, we hope to construct ball-like decision boundaries in the deep feature space for open intent detection. 

Due to the specificity of different intent categories, we aim to learn the corresponding decision boundaries for each known class. Concretely, for the $k^{\text{th}}$ known class, the spherical decision boundary is determined by its corresponding centroid $\boldsymbol{c}_{k}$ and radius $\Delta_{k}$, where $k \in \{1,2,\cdots, K\}$.

The centroid $\boldsymbol{c}_{k}$ is the average intent representation in class $k$, as defined in section~\ref{centers_def}. As the decision boundaries need to be adaptive to the intent feature space, the radius should be learnable to control the space range of the closed ball area. For this purpose, $\Delta_{k}$ is learned by the neural network with a 
boundary parameter $\widehat{\Delta_{k}} \in \mathds R$ . As suggested in~\cite{tapaswi2019video}, the $\operatorname{Softplus}$ activation function is utilized as the mapping between $\Delta_{k}$ and $\widehat{\Delta_{k}}$:
\begin{align}
\Delta_{k}=\log \left(1+\exp(\widehat{\Delta_{k}})\right).
\label{delta}
\end{align}

The $\operatorname{Softplus}$ activation function is selected for the following reasons. Firstly, it guarantees differentiability with different $\widehat{\Delta_{k}} \in \mathds R$ and supports stable optimization. Secondly, it ensures the radius $\Delta_{k}$ is above zero. Finally, it achieves linear characteristics like $\operatorname{ReLU}$ and allows for bigger $\Delta_{k}$ if necessary.

\subsubsection{Boundary Learning}
After formulating the centroid $\boldsymbol{c}_{k}$ and the radius $\Delta_{k}$, a critical problem is how to find the suitable decision boundary with the prior knowledge of only known intent feature distributions. For each known class, a tight decision boundary should balance both empirical and open space risks~\cite{scheirer2013toward}. That is, a tradeoff between both inside and outside examples belonging to its class is required.

For each example $(\boldsymbol{z}_i, y_{i})$, if $\|\boldsymbol{z}_{i}-\boldsymbol{c}_{y_{i}}\|_{2} > \Delta_{y_{i}}$,
the decision boundaries are too small to contain their corresponding known intent examples, which may increase the empirical risk. In contrast, if $\|\boldsymbol{z}_{i}-\boldsymbol{c}_{y_{i}}\|_{2} < \Delta_{y_{i}}$, though larger decision boundaries are beneficial to identify more known intent examples, they are more likely to introduce more open intent examples, which may increase the open space risk. Thus, the boundary loss $\mathcal{L}_{b}$ is proposed to make a tradeoff:
\begin{equation}
\begin{split}
\mathcal{L}_{b}=\frac{1}{N}&\sum_{i=1}^{N}\left[\delta_{i}\left(\|\boldsymbol{z}_{i}-\boldsymbol{c}_{y_{i}}\|_{2}
-\Delta_{y_{i}}\right)\right.\\
&+\left. \left(1-\delta_{i}\right) \left(\Delta_{y_{i}}-\|\boldsymbol{z}_{i}-\boldsymbol{c}_{y_{i}}\|_{2}
\right)\right],
\end{split} 
\end{equation}
where $\delta_{i}$ is defined as:
\begin{equation}
\delta_{i}:=\left\{\begin{array}{ll}1, \text { if } & \|\boldsymbol{z}_{i}-\boldsymbol{c}_{y_{i}}\|_{2} > \Delta_{y_{i}}, \\ 0, \text { if } & \|\boldsymbol{z}_{i}-\boldsymbol{c}_{y_{i}}\|_{2} \leq \Delta_{y_{i}}.\end{array}\right.
\end{equation}
The boundary parameter $\widehat{\Delta_{k}}$ is updated regarding to $\mathcal{L}_{b}$ as follows:
\begin{equation}
\widehat{\Delta_{k}} :=\widehat{\Delta_{k}}-\eta \frac{\partial \mathcal{L}_{b}}{\partial \widehat{\Delta_{k}}},
\end{equation} 
where $\eta$ is the learning rate and $\frac{\partial \mathcal{L}_{b}}{\partial \widehat{\Delta_{k}}}$ is computed by:
\begin{equation}	
\frac{\partial \mathcal{L}_{b}}{\partial \widehat{\Delta_{k}}}=\frac{\sum_{i=1}^{N} \delta^{'}\left(y_{i}=k\right) \cdot(-1)^{\delta_{i}}}{\sum_{i=1}^{N} \delta^{'}\left(y_{i}=k\right)}\cdot \frac{1}{1+e^{-\widehat{\Delta_{k}}}},
\end{equation}
where $\delta^{'}(y_{i}=k)=1$ if $y_{i}=k$ and $\delta^{'}(y_{i}=k)=0$ if not. The denominator is guaranteed to be not zero by  updating only $\widehat{\Delta_{k}}$ that has examples belonging to class $k$ in a mini-batch. 

The intrinsic properties of the boundary loss $\mathcal{L}_{b}$ are in favor of learning adaptive decision boundaries. It calculates the Euclidean distance between each example and its centroid $\|\boldsymbol{z}_{i}-\boldsymbol{c}_{y_{i}}\|_{2}$ and uses the distance to compare with the radius of the corresponding decision boundary $\Delta_{y_{i}}$, yielding the inside loss or outside loss. Specifically, the inside (or outside) loss is computed by the sum of the distances between the $k^{\text{th}}$ class boundary-inside (or boundary-outside) examples and the boundary $\Delta_{k}$. The suitable decision boundary is a balance between the inside and outside losses. When the inside loss is larger, the cumulative gradients are positive to move inward the decision boundary. Similarly, when the outside loss is larger, the cumulative gradients are negative to make the decision boundary expand outward. This process enables the decision boundaries to be adaptive to the known intent feature space until balance.
\subsection{Open Classification with Decision Boundary}
After boundary learning, the learned decision boundaries and centroids are used for inference. For each example $\boldsymbol{z}_{i}$, it is first recognized as the index $k_{a}$ of the nearest centroid. Then, the corresponding decision boundary $\Delta_{k_{a}}$ is utilized to detect whether it belongs to the open intent:
\begin{equation}	k_{a}=\underset{k}{\operatorname{argmin}}\left\{d(\boldsymbol{z}_{i},\boldsymbol{c}_{k})\right\}_{k \in I^{\text{Known}}},
\end{equation}
\begin{equation}
\hat{y}=\left\{\begin{array}{l}
\text {\textit{Open}, if } d(\boldsymbol{z}_{i},\boldsymbol{c}_{k}) > \Delta_{k_{a}}; \\
k_{a}, \text {   if } d(\boldsymbol{z}_{i},\boldsymbol{c}_{k}) \leq \Delta_{k_{a}},
\end{array}\right.
\end{equation}
where $I^{known}$ is the known intent label set as mentioned in section~\ref{problem_formulate} and $d(\boldsymbol{z}_{i},\boldsymbol{c}_{k})$ denotes the Euclidean distance between $\boldsymbol{z}_{i}$ and $\boldsymbol{c}_{k}$. 
\section{Experiments}
This section introduces the benchmark datasets, baselines,
evaluation metrics, experimental settings, and results.
\subsection{Datasets}
We conduct experiments on three challenging real-world datasets to evaluate our approach. The detailed statistics are shown in Table~\ref{data-stat-table}. 

BANKING: A fine-grained dataset in the banking domain~\cite{Casanueva2020}. It contains 77 intents and 13,083 customer service queries. We split a validation set of 1,000 samples from the original training set. 

OOS: A dataset for intent classification and out-of-scope prediction~\cite {larson-etal-2019-evaluation}. It contains 150 intents, 22,500 in-domain queries and 1,200 out-of-domain queries.

StackOverflow: The dataset was published on Kaggle.com. It contains 3,370,528 technical question titles. We use the processed dataset~\cite {xu2015short}, which has 20 different classes and 1,000 samples for each class.

\begin{table*}[t!]\small
	\caption{ \label{data-stat-table} Statistics of BANKING, OOS and StackOverflow datasets. \# indicates the total number of utterances.}
	\centering
	\begin{tabular}{@{\extracolsep{15pt}} ccccccc @{}}
		\toprule
		Dataset & Classes & \#Training & \#Validation & \#Test & Vocabulary Size & Length (max / mean) \\
		\midrule
		BANKING & 77 & 9,003 & 1,000 & 3,080 & 5,028 & 79 / 11.91\\
		OOS & 150 & 15,000 & 3,000 & 5,700 & 8,376 & 28 / 8.31 \\
		StackOverflow & 20 & 12,000 & 2,000 & 6,000 & 17,182 & 41 / 9.18 \\
		\bottomrule
	\end{tabular}
\end{table*}
\subsection{Baselines}
We compare our approach with state-of-the-art methods in open set recognition: OpenMax~\cite{bendale2016towards}, DOC~\cite{Shu2017DOCDO}, ARPL~\cite{9521769} and open intent detection: DeepUnk~\cite{lin-xu-2019-deep}, SEG~\cite{yan-etal-2020-unknown}, ($K$+1)-way~\cite{zhan-etal-2021-scope}, ADB~\cite{Zhang_Xu_Lin_2021}. Besides, three OOD detection baselines are built for this task: MSP~\cite{hendrycks17baseline}, LOF~\cite{breunig2000lof}, MDF~\cite{xu2021unsupervised}. The detailed information is as follows:

\subsubsection{MSP} It is a simple baseline that predicts known classes with the maximum softmax probabilities and rejects the negative samples with the threshold of 0.5. 
\subsubsection{LOF} It is a density-based method to detect the low-density outliers as the open-class samples. The hyper-parameters are set as in DeepUnk~\cite{lin-xu-2019-deep}.
\subsubsection{MDF} It is an unsupervised OOD detection method adapted to our task. We first train the model on ID data by applying both cross-entropy and MLM losses. Once the model is well-trained, we utilize it to identify known intents and extract averaged representations from each pre-trained transformer layer. These representations are then used to calculate Mahalanobis distances, which are concatenated as features. Finally, the features are fed into a one-class SVM to detect OOD samples. For better performance, we choose the RBF kernel and set the lower bound on the fraction of support vectors to 0.1.
\subsubsection{OpenMax} It first uses the softmax loss to train a classifier on the known intents and then fits a Weibull distribution to the classifier's output logits. The confidence scores are finally calibrated with the OpenMax Layer. The default hyper-parameters in ~\cite{bendale2016towards} are adopted (Weibull tail size is 20).
\subsubsection{DOC} It rejects the open class by calculating different probability thresholds of each known class with Gaussian fitting. The default hyper-parameters in~\cite{Shu2017DOCDO} are adopted (the number of standard deviations away from the mean is 3).
\subsubsection{ARPL} It learns representations of reciprocal points by maximizing the variance between them and known-class samples. It combines Euclidean and cosine distances as the metric and sets a learnable margin to constrain the open space. However, the performance collapses after directly applying the approach in computer vision to our task. Thus, we alleviate the issue by pre-training with known intents in advance. The probability threshold is set to 0.5 for detecting the open class.
\subsubsection{DeepUnk} It first uses the margin loss to learn deep features and then detects the unknown class with LOF. The cosine margin and scaling factor are set to 0.35 and 30, respectively. 
\subsubsection{SEG} It incorporates the semantic information of each class into the large margin Gaussian mixture loss~\cite{8579048} for feature representation, followed by a LOF detector.  
\subsubsection{(K+1)-way} It constructs OOD samples by creating convex combinations of intent representations from two different ID classes, which are then treated as the ($K$+1)$^{\text{th}}$ open class data. The ID and OOD data are jointly trained using soft labels. The temperature parameter is set to 0.1 as suggested in~\cite{zhan-etal-2021-scope}.
\subsubsection{ADB} It is a variant of DA-ADB, which also learns adaptive decision boundaries based on the known intent feature space. The difference is that it uses softmax loss rather than leveraging the distance-aware concepts during pre-training.  

In particular, all methods are used the same BERT model as backbones for a fair comparison.

\subsection{Evaluation Metrics}
For open intent detection, the accuracy score (ACC) and the macro F1-score over all classes (F1) are used to evaluate the overall performance. The macro F1-score over known classes ($\text{F1}_{\text{known}}$) and over the open class ($\text{F1}_{\text{open}}$)  are used to evaluate the fine-grained performance. 


Given a set of classes $C=\{C_{1}, \cdots, C_{K}, C_{K\text{+1}}\}$, where $K$ is the number of known classes and $C_{K\text{+1}}$ is the open class. The macro F1-score over all classes (F1) is computed by:
\begin{align}
\text{F1} &= 2 \times  \frac{\text{P} \times \text{R}}{\text{P} + \text{R}},
\end{align}
\begin{align}
\text{P}=\frac{\sum_{i=1}^{K+1} \text{P}_{C_{i}}
}{K+1},\quad 
\text{R}=\frac{\sum_{i=1}^{K+1} \text{R}_{C_{i}}
}{K+1},
\end{align}
\begin{align}
\label{P+R}
\text{P}_{C_{i}}=\frac{\text{TP}_{C_{i}}}{\text{TP}_{C_{i}}+\text{FP}_{C_{i}}},\quad \text{R}_{C_{i}}=\frac{\text{TP}_{C_{i}}}{\text{TP}_{C_{i}}+\text{FN}_{C_{i}}},
\end{align}
where $\text{P}$ and $\text{R}$ are the macro precision score and the macro recall score over $K$+1 classes. $\text{P}_{C_{i}}$ and $\text{R}_{C_{i}}$ are the precision score and recall score on the $C_{i}$ class. $\text{TP}_{C_{i}}$,  $\text{FP}_{C_{i}}$  and $\text{FN}_{C_{i}}$ are the true positives, false positives and false negatives of the $C_{i}$ class, respectively.  
Similarly, the macro F1-score over known classes ($\text{F1}_{\text{known}}$) and the open class ($\text{F1}_{\text{open}}$) are computed by:
\begin{align}
\text{F1}_{\text{known}} &= 2 \times  \frac{\text{P}_{\text{known}} \times \text{R}_{\text{known}}}{\text{P}_{\text{known}} + \text{R}_{\text{known}}},
\end{align}
\begin{align}
\text{P}_{\text{known}}=\frac{\sum_{i=1}^{K} \text{P}_{C_{i}}
}{K},\quad 
\text{R}_{\text{known}}=\frac{\sum_{i=1}^{K} \text{R}_{C_{i}}
}{K},
\end{align}
\begin{align}
\text{F1}_{\text{open}} &= 2 \times  \frac{\text{P}_{C_{K+1}} \times \text{R}_{C_{K+1}}}{\text{P}_{C_{K+1}} + \text{R}_{C_{K+1}}},
\end{align}
where $\text{P}_{C_{i}}$ and $\text{R}_{C_{i}}$ are computed the same as in Eq.~\ref{P+R}.

\begin{table*}[t!]\small
	\caption{ \label{results-main-1}  
		Overall performance of open intent detection with different known class ratios (25\%, 50\% and 75\%) and their mean scores on three datasets. The proposed method DA-ADB and its variant ADB are significantly better than others with  $p$-value $<$ 0.05 ($\dagger$) and $p$-value $<$ 0.1 (*) using t-test.
	}
	\centering
	\begin{tabular}{@{\extracolsep{8pt}}c|cccccccccc}
		\toprule
		\multirow{2}{*}{Datasets} 
		& \multirow{2}{*}{Methods} & \multicolumn{2}{c}{25\%} & \multicolumn{2}{c}{50\%} & \multicolumn{2}{c}{75\%} &
		\multicolumn{2}{c}{Mean} \\ \cline{3-4}\cline{5-6}\cline{7-8}\cline{9-10} \addlinespace[0.1cm]
		& & ACC & F1  & ACC & F1 & ACC & F1 & ACC & F1 \\ 
		\hline\addlinespace[0.1cm]
		\multirow{9}{*}{BANKING} 
		
		& MSP & 42.19$\dagger$ & 49.92$\dagger$ & 61.67$\dagger$ & 72.51$\dagger$ & 77.08$\dagger$ & 84.33$\dagger$ & 60.31$\dagger$ & 68.92$\dagger$ \\
		& SEG & 48.73$\dagger$ & 51.49$\dagger$ & 55.11$\dagger$ & 63.32$\dagger$ & 64.65$\dagger$ & 69.54$\dagger$ & 56.16$\dagger$ & 61.45$\dagger$ \\
		& OpenMax & 47.76$\dagger$ & 53.18$\dagger$ & 65.53$\dagger$ & 74.64$\dagger$ & 78.32$\dagger$ & 84.95$\dagger$ & 63.87$\dagger$ & 70.92$\dagger$ \\
        & MDF & 77.17$\dagger$ & 46.85$\dagger$ & 60.18$\dagger$ & 64.10$\dagger$ & 64.59$\dagger$ & 74.76$\dagger$ & 67.31$\dagger$ & 61.90$\dagger$ \\
		& LOF & 66.73$\dagger$ & 63.38$\dagger$ & 71.13$\dagger$ & 76.26$\dagger$ & 77.21$\dagger$ & 83.64$\dagger$ & 71.69$\dagger$ & 74.43$\dagger$  \\
		& DOC & 70.31$\dagger$ & 65.74$\dagger$ & 74.60$\dagger$ & 78.24$\dagger$ & 78.94$\dagger$ & 83.79$\dagger$ & 74.62$\dagger$ & 75.92$\dagger$ \\
		& DeepUNK & 70.68$\dagger$ & 65.57$\dagger$ & 71.01$\dagger$ & 75.41$\dagger$ & 74.73$\dagger$ & 81.12$\dagger$ & 72.14$\dagger$ & 74.03$\dagger$ \\
		& ($K$+1)-way & 76.66$\dagger$ & 68.44$\dagger$ & 74.65$\dagger$ & 77.83$\dagger$ & 79.18$\dagger$ & 84.71$\dagger$ & 76.38$\dagger$ & 76.99$\dagger$ \\
		& ARPL & 76.80$\dagger$ & 64.01$\dagger$ & 74.11$\dagger$ & 77.77$\dagger$ & 79.60$\dagger$ & 85.16$\dagger$ & 76.84$\dagger$ & 75.65$\dagger$ \\
		\cline{2-10} \addlinespace[0.1cm]
		& ADB & 79.33 & 71.63 & 79.61$\dagger$ & 81.34$\dagger$ & \textbf{81.39} & \textbf{86.11} & 80.11* & 79.69 \\
		& DA-ADB & \textbf{81.19} & \textbf{73.73} & \textbf{81.51} & \textbf{82.53} & 81.12 & 85.65 & \textbf{81.27} & \textbf{80.64}\\
		\midrule
		\midrule 
		\multirow{9}{*}{OOS} 
		
		& MSP & 53.38$\dagger$ & 51.23$\dagger$ & 66.68$\dagger$ & 72.70$\dagger$ & 76.19$\dagger$ & 83.48$\dagger$ & 65.42$\dagger$ & 69.14$\dagger$ \\
		& SEG & 52.18$\dagger$ & 47.00$\dagger$ & 60.67$\dagger$ & 62.55$\dagger$ & 42.78$\dagger$ & 42.70$\dagger$ & 51.88$\dagger$ & 50.75$\dagger$ \\
		& OpenMax & 70.27$\dagger$ & 63.03$\dagger$ & 80.22$\dagger$ & 79.86$\dagger$ & 75.36$\dagger$ & 71.17$\dagger$ & 75.28$\dagger$ & 71.35$\dagger$ \\
        & MDF & 76.56$\dagger$ & 50.34$\dagger$ & 60.72$\dagger$ & 61.61$\dagger$ & 63.98$\dagger$ & 72.02$\dagger$ & 67.09$\dagger$ & 61.32$\dagger$ \\
		& LOF & 87.77 & 78.13 & 85.22$\dagger$ & 83.86$\dagger$ & 85.07$\dagger$ & 87.20$\dagger$ & 86.02$\dagger$ & 83.06$\dagger$ \\
		& DOC & 86.08$\dagger$ & 75.86$\dagger$ & 85.19$\dagger$ & 83.89$\dagger$ & 85.93$\dagger$ & 87.87$\dagger$ & 85.73$\dagger$ & 82.54$\dagger$ \\
		& DeepUNK & 87.18 & 77.32* & 84.95$\dagger$ & 83.35$\dagger$ & 84.61$\dagger$ & 86.53$\dagger$ & 85.58$\dagger$ & 82.40$\dagger$ \\
		& ($K$+1)-way & 85.36$\dagger$ & 74.43$\dagger$ & 82.19$\dagger$ & 81.56$\dagger$ & 83.51$\dagger$ & 86.66$\dagger$ & 83.69$\dagger$ & 80.88$\dagger$ \\
		& ARPL & 84.51$\dagger$ & 73.44$\dagger$ & 80.36$\dagger$ & 80.88$\dagger$ & 81.29$\dagger$ & 86.00$\dagger$ & 82.05$\dagger$ & 80.11$\dagger$ \\
		\cline{2-10} \addlinespace[0.1cm]
		& ADB & 88.30 & 78.23 & 86.54$\dagger$ & 85.16 & 86.99 & \textbf{88.94} & 87.28$\dagger$ & 84.11 \\
		& DA-ADB & \textbf{89.48} & \textbf{79.92} & \textbf{87.93} & \textbf{85.64} & \textbf{87.39} & 88.41 & \textbf{88.27} & \textbf{84.66} \\
		\midrule
		\midrule
		\multirow{9}{*}{StackOverflow} 
		
		& MSP & 27.91$\dagger$ & 37.49$\dagger$ & 53.23$\dagger$ & 62.70$\dagger$ & 73.20$\dagger$ & 78.70$\dagger$ & 51.45$\dagger$ & 59.63$\dagger$ \\
		& SEG & 23.33$\dagger$ & 34.40$\dagger$ & 43.04$\dagger$ & 55.10$\dagger$ & 62.72$\dagger$ & 69.97$\dagger$ & 43.03$\dagger$ & 53.16$\dagger$ \\
		& OpenMax & 38.97$\dagger$ & 45.35$\dagger$ & 60.27$\dagger$ & 67.72$\dagger$ & 75.78$\dagger$ & 80.90$\dagger$ & 58.34$\dagger$ & 64.66$\dagger$ \\
        & MDF & 74.10$\dagger$ & 53.95$\dagger$ & 56.46$\dagger$ & 61.47$\dagger$ &62.98$\dagger$ & 71.12$\dagger$ & 64.51$\dagger$ & 62.18$\dagger$ \\
		& LOF & 25.02$\dagger$ & 35.29$\dagger$ & 44.56$\dagger$ & 56.57$\dagger$ & 65.05$\dagger$ & 71.87$\dagger$ & 44.88$\dagger$ & 54.58$\dagger$  \\
		& DOC & 57.75$\dagger$ & 57.34$\dagger$ & 73.88$\dagger$ & 76.80$\dagger$ & 80.55$\dagger$ & 84.37$\dagger$ & 70.73$\dagger$ & 72.84$\dagger$ \\
		& DeepUNK & 40.03$\dagger$ & 45.64$\dagger$ & 55.46$\dagger$ & 64.78$\dagger$ & 71.56$\dagger$ & 77.63$\dagger$ & 55.68$\dagger$ & 62.68$\dagger$ \\
		& ($K$+1)-way & 49.75$\dagger$ & 50.82$\dagger$ & 62.57$\dagger$ & 68.81$\dagger$ & 74.00$\dagger$ & 78.95$\dagger$ & 62.11$\dagger$ & 66.19$\dagger$ \\
		& ARPL & 66.76$\dagger$ & 62.62$\dagger$ & 75.65$\dagger$ & 77.87$\dagger$ & 79.64$\dagger$ & 83.85$\dagger$ & 74.02$\dagger$ & 74.78$\dagger$ \\
		\cline{2-10} \addlinespace[0.1cm]
		& ADB & 86.75 & 79.85$\dagger$ & 86.49$\dagger$ & 85.54$\dagger$ & 82.89 & 86.11* & 85.38$\dagger$ & 83.83$\dagger$ \\
		& DA-ADB & \textbf{89.07} & \textbf{82.83} & \textbf{87.78} & \textbf{86.91} & \textbf{83.56} & \textbf{86.84} & \textbf{86.80} & \textbf{85.53} \\
		\bottomrule 
	\end{tabular}
\end{table*}

\begin{table*}[t!]\small
	\caption{ \label{results-main-2}  
		Fine-grained performance of open intent detection with different known class ratios (25\%, 50\%, 75\%) and their mean scores on three datasets. The proposed method DA-ADB and its variant ADB are significantly better than others with  $p$-value $<$ 0.05 ($\dagger$) and $p$-value $<$ 0.1 (*) using t-test.}
	\centering
	\begin{tabular}{@{\extracolsep{8pt}}c|cccccccccc}
		\toprule
		\multirow{2}{*}{Datasets} 
		& \multirow{2}{*}{Methods} & \multicolumn{2}{c}{25\%} & \multicolumn{2}{c}{50\%} & \multicolumn{2}{c}{75\%} &
		\multicolumn{2}{c}{Mean} \\ \cline{3-4}\cline{5-6}\cline{7-8}\cline{9-10} \addlinespace[0.1cm]
		& & Open & Known  & Open & Known & Open & Known & Open & Known \\ 
		\hline\addlinespace[0.1cm]
		\multirow{9}{*}{BANKING} 
		
		& MSP & 39.42$\dagger$ & 50.47$\dagger$ & 46.29$\dagger$ & 73.20$\dagger$ & 46.05$\dagger$ & 84.99$\dagger$ & 43.92$\dagger$ & 69.55$\dagger$ \\
		& SEG & 51.58$\dagger$ & 51.48$\dagger$ & 43.03$\dagger$ & 63.85$\dagger$ & 37.22$\dagger$ & 70.10$\dagger$ & 43.94$\dagger$ & 61.81$\dagger$ \\
		& OpenMax & 48.52$\dagger$ & 53.42$\dagger$ & 55.03$\dagger$ & 75.16$\dagger$ & 53.02$\dagger$ & 85.50$\dagger$ & 52.19$\dagger$ & 71.36$\dagger$ \\
        & MDF & 85.70$\dagger$ & 44.80$\dagger$ & 57.72$\dagger$ & 64.27$\dagger$ & 33.43$\dagger$ & 75.47$\dagger$ & 58.95$\dagger$ & 61.51$\dagger$ \\
		& LOF & 72.64$\dagger$ & 62.89$\dagger$ & 66.81$\dagger$ & 76.51$\dagger$ & 54.19$\dagger$ & 84.15$\dagger$ & 64.55$\dagger$ & 74.52$\dagger$  \\
		& DOC & 76.64$\dagger$ & 65.16$\dagger$ & 72.66$\dagger$ & 78.38$\dagger$ & 63.51$\dagger$ & 84.14$\dagger$ & 70.94$\dagger$ & 75.89$\dagger$ \\
		& DeepUNK & 76.98$\dagger$ & 64.97$\dagger$ & 67.80$\dagger$ & 75.61$\dagger$ & 50.57$\dagger$ & 81.65$\dagger$ & 65.12$\dagger$ & 74.08$\dagger$ \\
		& ($K$+1)-way & 82.66$\dagger$ & 67.70$\dagger$ & 72.58$\dagger$ & 77.97$\dagger$ & 59.89$\dagger$ & 85.14 & 71.71$\dagger$ & 76.94$\dagger$ \\
		& ARPL & 83.39$\dagger$ & 62.99$\dagger$ & 71.79$\dagger$ & 77.93$\dagger$ & 61.26$\dagger$ & 85.58$\dagger$ & 72.15$\dagger$ & 75.50$\dagger$ \\
		\cline{2-10} \addlinespace[0.1cm]
		& ADB & 85.05 & 70.92 & 79.43$\dagger$ & 81.39* & 67.34* & \textbf{86.44} & 77.27$\dagger$ & 79.58 \\
		& DA-ADB & \textbf{86.57} & \textbf{73.05} & \textbf{81.93} & \textbf{82.54} & \textbf{69.37} & 85.93 & \textbf{79.29} & \textbf{80.51} \\
		\midrule
		\midrule
		\multirow{9}{*}{OOS} 
		
		& MSP & 59.26$\dagger$ & 51.02$\dagger$ & 63.71$\dagger$ & 72.82$\dagger$ & 63.86$\dagger$ & 83.65$\dagger$ & 62.28$\dagger$ & 69.16$\dagger$ \\
		& SEG & 59.22$\dagger$ & 46.67$\dagger$ & 61.34$\dagger$ & 62.57$\dagger$ & 40.74$\dagger$ & 42.72$\dagger$ & 53.77$\dagger$ & 50.65$\dagger$ \\
		& OpenMax & 77.51$\dagger$ & 62.65$\dagger$ & 82.15$\dagger$ & 79.83$\dagger$ & 75.18$\dagger$ & 71.14$\dagger$ & 78.28$\dagger$ & 71.21$\dagger$ \\
        & MDF & 84.89$\dagger$ & 49.43$\dagger$ & 62.31$\dagger$ & 61.60$\dagger$ & 51.33$\dagger$ & 72.21$\dagger$ & 66.18$\dagger$ & 61.08$\dagger$ \\
		& LOF & 91.96 & 77.77 & 87.57$\dagger$ & 83.81$\dagger$ & 82.81$\dagger$ & 87.24$\dagger$ & 87.45$\dagger$ & 82.94$\dagger$ \\
		& DOC & 90.78$\dagger$ & 75.46$\dagger$ & 87.45$\dagger$ & 83.84$\dagger$ & 83.87$\dagger$ & 87.91$\dagger$ & 87.37$\dagger$ & 82.40$\dagger$ \\
		& DeepUNK & 91.61 & 76.95* & 87.48$\dagger$ & 83.30$\dagger$ & 82.67$\dagger$ & 86.57$\dagger$ & 87.25$\dagger$ & 82.27$\dagger$ \\
		& ($K$+1)-way & 90.27$\dagger$ & 74.02$\dagger$ & 84.25$\dagger$ & 81.52$\dagger$ & 79.59$\dagger$ & 86.72* & 84.70$\dagger$ & 80.75$\dagger$ \\
		& ARPL & 89.63$\dagger$ & 73.01$\dagger$ & 81.81$\dagger$ & 80.87$\dagger$ & 74.67$\dagger$ & 86.10$\dagger$ & 812.04$\dagger$ & 79.99$\dagger$ \\
		\cline{2-10} \addlinespace[0.1cm]
		& ADB & 92.36 & 77.85 & 88.60$\dagger$ & 85.12 & 84.85$\dagger$ & \textbf{88.97} & 88.60$\dagger$ & 83.98 \\
		& DA-ADB & \textbf{93.20} & \textbf{79.57} & \textbf{90.10} & \textbf{85.58} & \textbf{86.00} & 88.43 & \textbf{89.77} & \textbf{84.53} \\
		\midrule
		\midrule
		\multirow{9}{*}{StackOverflow} 
		
		& MSP & 11.66$\dagger$ & 42.66$\dagger$ & 26.94$\dagger$ & 66.28$\dagger$ & 37.86$\dagger$ & 81.42$\dagger$ & 25.49$\dagger$ & 63.45$\dagger$ \\
		& SEG & 4.19$\dagger$ & 40.44$\dagger$ & 4.72$\dagger$ & 60.14$\dagger$ & 6.00$\dagger$ & 74.24$\dagger$ & 4.97$\dagger$ & 58.27$\dagger$ \\
		& OpenMax & 34.52$\dagger$ & 47.51$\dagger$ & 46.11$\dagger$ & 69.88$\dagger$ & 49.69$\dagger$ & 82.98$\dagger$ & 43.44$\dagger$ & 66.79$\dagger$ \\
        & MDF & 83.03$\dagger$ & 48.13$\dagger$ & 50.19$\dagger$ & 62.60$\dagger$ & 28.52$\dagger$ & 73.96$\dagger$ & 53.91$\dagger$ & 61.56$\dagger$ \\
		& LOF & 7.14$\dagger$ & 40.92$\dagger$ & 5.18$\dagger$ & 61.71$\dagger$ & 5.22$\dagger$ & 76.31$\dagger$ & 5.85$\dagger$ & 59.65$\dagger$  \\
		& DOC & 62.50$\dagger$ & 56.30$\dagger$ & 71.18$\dagger$ & 77.37$\dagger$ & 65.32$\dagger$ & 85.64$\dagger$ & 66.33$\dagger$ & 73.10$\dagger$ \\
		& DeepUNK & 36.87$\dagger$ & 47.39$\dagger$ & 35.80$\dagger$ & 67.67$\dagger$ & 34.38$\dagger$ & 77.63$\dagger$ & 35.68$\dagger$ & 65.19$\dagger$ \\
		& ($K$+1)-way & 52.23$\dagger$ & 50.54$\dagger$ & 51.69$\dagger$ & 70.53$\dagger$ & 45.22$\dagger$ & 81.20$\dagger$ & 49.71$\dagger$ & 67.42$\dagger$ \\
		& ARPL & 72.95$\dagger$ & 60.55$\dagger$ & 73.97$\dagger$ & 78.26$\dagger$ & 62.99$\dagger$ & 85.24$\dagger$ & 69.97$\dagger$ & 74.68$\dagger$ \\
		\cline{2-10} \addlinespace[0.1cm]
		& ADB & 90.96 & 77.62$\dagger$ & 87.70$\dagger$ & 85.32$\dagger$ & 74.10 & 86.91* & 84.25$\dagger$ & 83.28$\dagger$ \\
		& DA-ADB & \textbf{92.65} & \textbf{80.87} & \textbf{88.86} & \textbf{86.71} & \textbf{74.55} & \textbf{87.66} & \textbf{85.35} & \textbf{85.08} \\
		\bottomrule 
	\end{tabular}
\end{table*}

\subsection{Experimental Settings}
Open intent detection follows the open-world setting~\cite{Shu2017DOCDO},  which keeps some classes as unknown (open) and integrates them back during testing. Specifically, the proportions of known classes to total categories are varied with 25\%, 50\%, and 75\%. The remaining classes are regarded as one open class. All the datasets are divided into training, validation, and testing sets. The samples from the open class are removed from the training and evaluation sets and only exist in the testing set, as mentioned in section~\ref{problem_formulate}. 
To reduce the impact of different selected known intent categories on the performance, we report the average performance over ten runs of experiments for each known class ratio with random seeds of 0-9.

We employ the pre-trained language model BERT (bert-uncased, with 12-layer transformer) implemented in PyTorch~\cite{wolf-etal-2020-transformers} and adopt most of its suggested hyper-parameters for optimization. To improve the training efficiency and achieve better performance, we freeze all but the last transformer layer parameters of BERT. The feature dimension $D$ is 768, the training batch size is 128, and the learning rate is 2e-5. For DA-ADB, the scalar $\alpha$ of the cosine classifier is 4, which is searched from $\{
2, 4, 8, 16, 32, 64\}$ by combining both feature learning and open classification performance on the evaluation set. The boundary loss $\mathcal{L}_{b}$ uses Adam~\cite {kingma2014adam} to optimize the boundary parameters at a learning rate of 0.05.

\subsection{Results}

The main experimental results of open intent detection are presented in Table~\ref{results-main-1} and Table~\ref{results-main-2}. The best results are highlighted in bold, and Student's t-test is conducted to measure the significance of performance difference between the best-performing method and other methods for each evaluation metric.  

Table~\ref{results-main-1} shows the overall performance of the accuracy score (ACC) and macro F1-score (F1) over all classes. The proposed approach DA-ADB and its variant ADB achieve the best results in all settings and outperform other baselines significantly. Compared with ADB,  DA-ADB yields substantial improvements with fewer known intents ($25\%$ and $50\%$) and achieves competitive results with more known intents ($75\%$), which indicates the effectiveness of the distance-aware concept for representation learning. Compared with the state-of-the-art methods, the average performance of three known-class ratios shows that DA-ADB improves ACC by 4.43\% on BANKING, 2.25\% on OOS, and 12.78\% on StackOverflow, respectively. 

Table~\ref{results-main-2} shows the fine-grained performance of the macro F1-score over known classes ($\text{F1}_{\text{known}}$) and the open class ($\text{F1}_{\text{open}}$). Our approach achieves significant improvements in detecting the open intent and largely enhances the known intent identification performance. We notice that ADB gains the best results over all baselines. It indicates that the learned decision boundaries are suitable to balance both the empirical and open space risks. On this basis, DA-ADB learns more friendly intent representations with distance information, which helps increase $1\%\sim2\%$ scores for open intent detection.

Moreover, it is worth noting that the improvements on the StackOverflow dataset are much more drastic than the other two datasets. We suppose the reason is that the characteristics of StackOverflow put forward higher requirements for open intent detection. Existing methods are limited to distinguishing the difficult semantic intents as technical question titles in StackOverflow without learning discriminative intent representations and decision boundaries. 
\begin{figure}
  \centering
  \subfigure[Vanilla intent representations.]{
    \includegraphics[width=0.465\columnwidth]{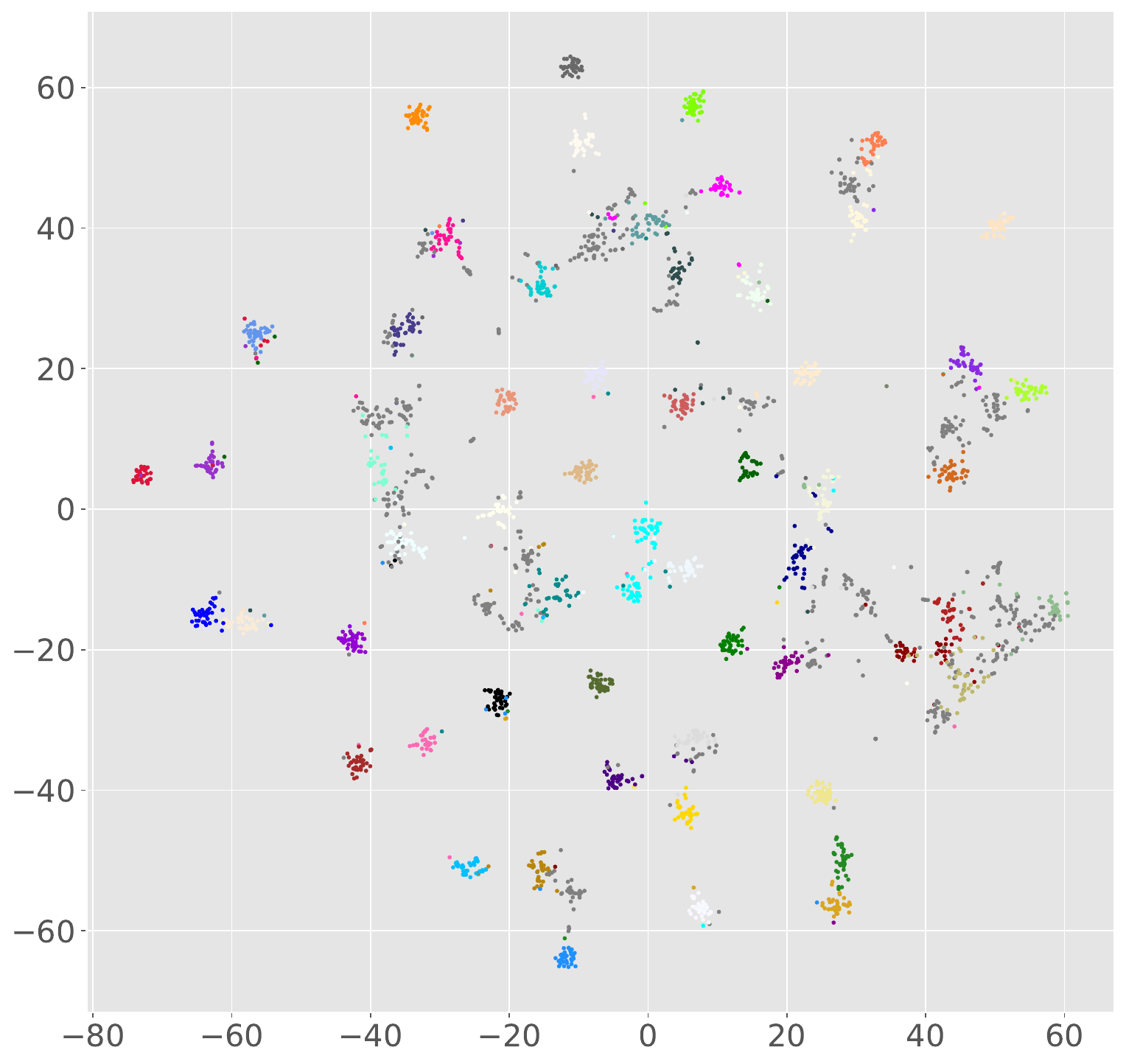}
  }
  \subfigure[Intent representations learned with distance-aware concepts.]{
    \includegraphics[width=0.465\columnwidth]{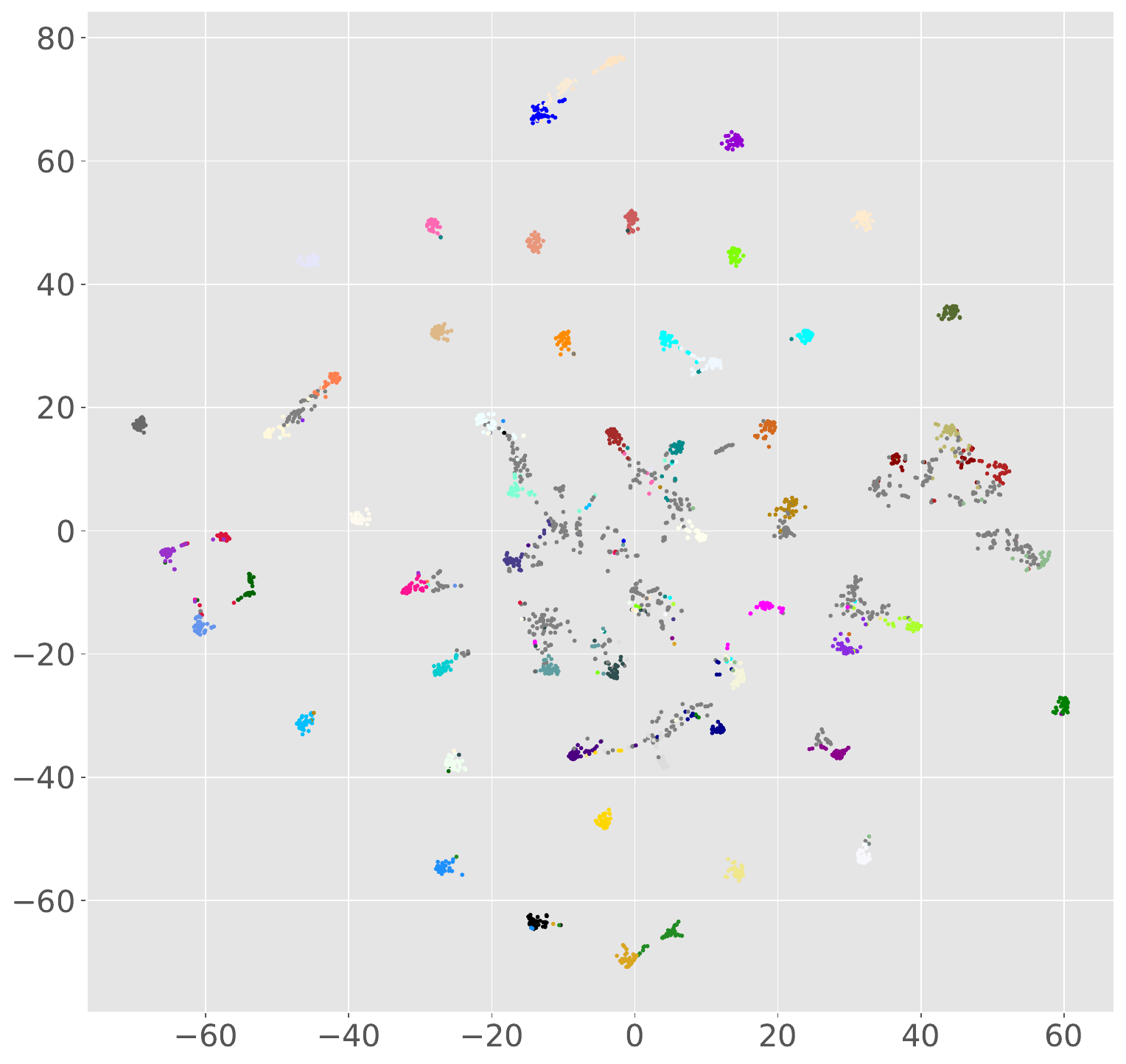}
  }
  \caption{Visualization of learned intent representations on the BANKING dataset. The open intent samples are marked in gray, and known intents are marked in other colors.}
  \label{feats_visualization}
\end{figure}

\begin{figure*}[!t]
	\centering  
	\includegraphics[width=2\columnwidth ]{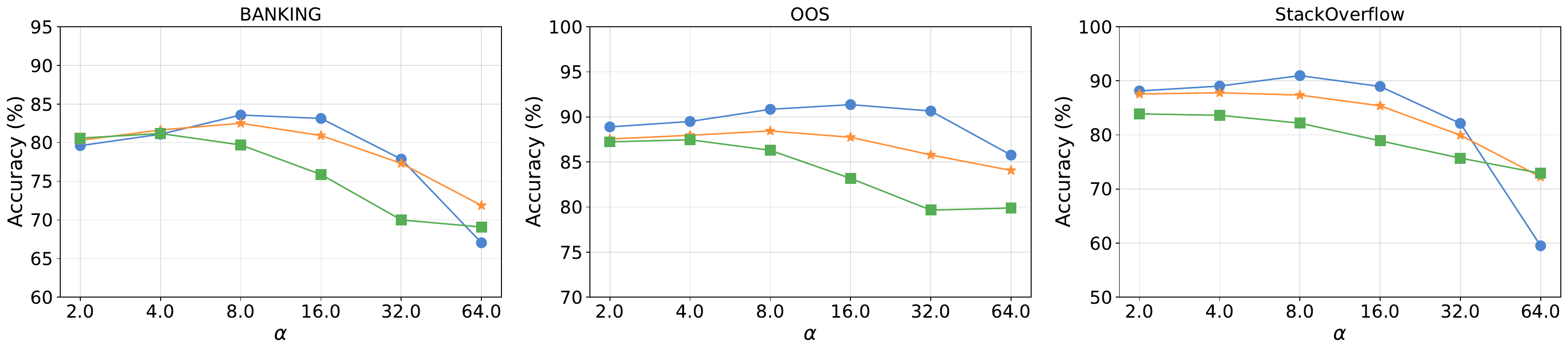}
	\includegraphics[width=1\columnwidth ]{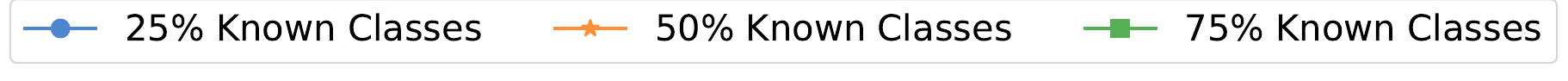}
	\caption{Influence of $\alpha$ on three datasets with different known class ratios.}
	\label{Aba_4}
\end{figure*}

\begin{figure*}[!t]
	\centering
	\includegraphics[width=2\columnwidth ]{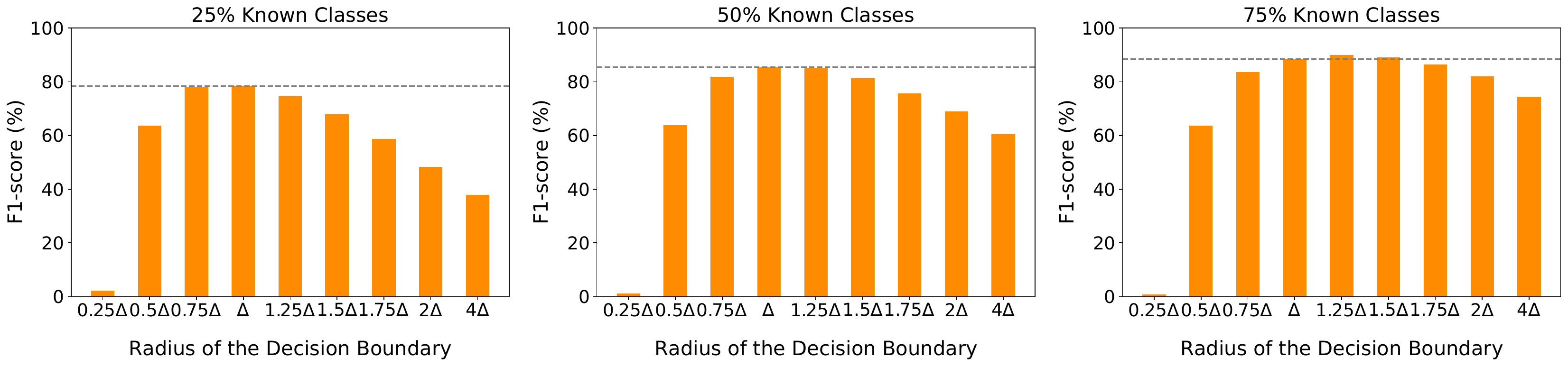}
	\caption{ Influence of the decision boundary on the OOS dataset with different known class ratios.} 
	\label{boundary}
\end{figure*}
\begin{figure}
	\centering  
	\includegraphics[width=0.8\columnwidth]{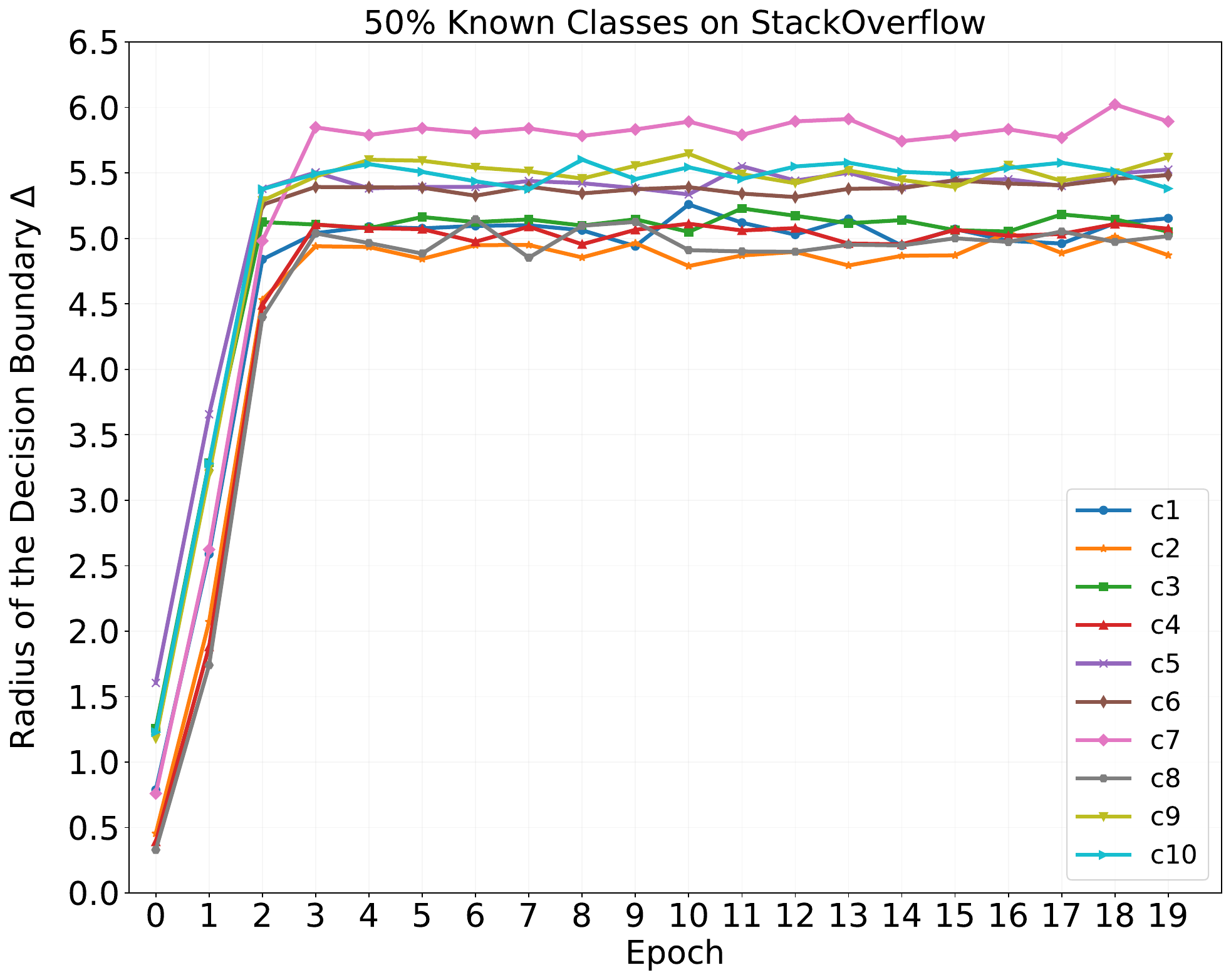}
	\caption{The boundary learning process.}
	\label{Aba_2_1}
\end{figure}
\section{Discussions}
This section investigates the effect of distance-aware representation, adaptive decision boundary learning, and labeled data. The first subsection visualizes the intent representations to verify the effectiveness of the distance-aware representation learning strategy and analyzes the influence of the hyper-parameter $\alpha$ mentioned in~\ref{alpha}. The second subsection compares the performance with different radii to demonstrate the compactness of the learned decision boundaries and visualizes the boundary learning process. The final subsection compares the robustness of different methods with less labeled data.

\begin{figure*}[!t]
	\centering  
	\includegraphics[width=2\columnwidth ]{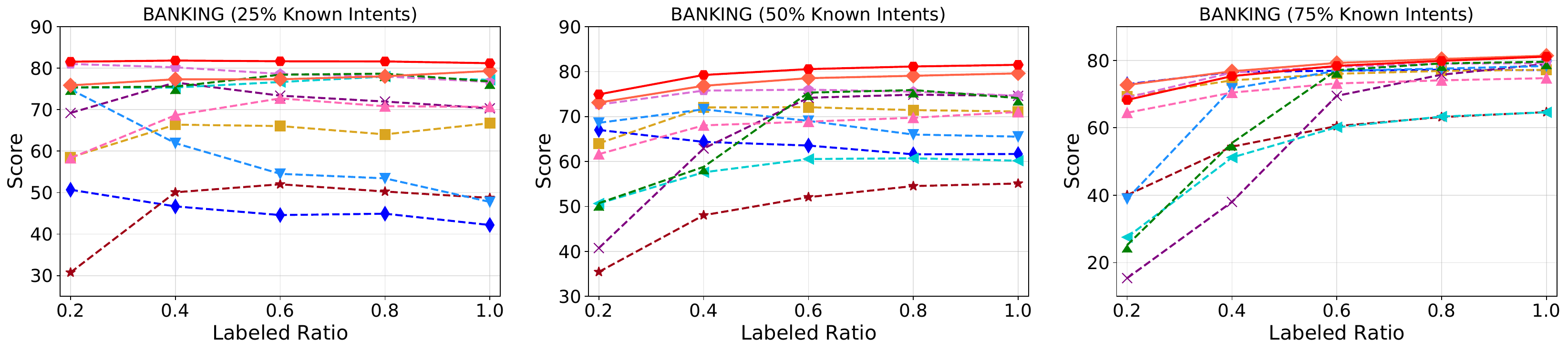}
	\includegraphics[width=2\columnwidth ]{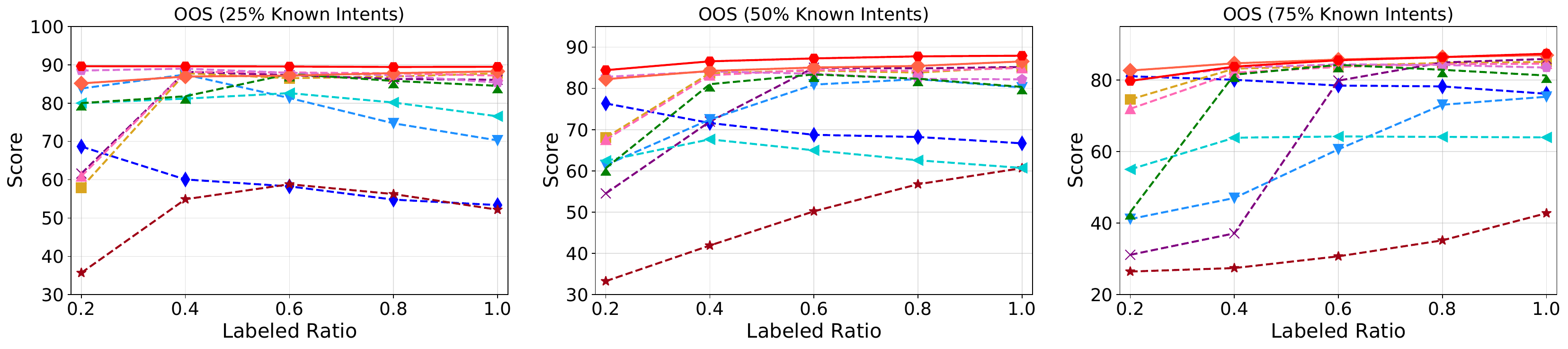}
	\includegraphics[width=2\columnwidth ]{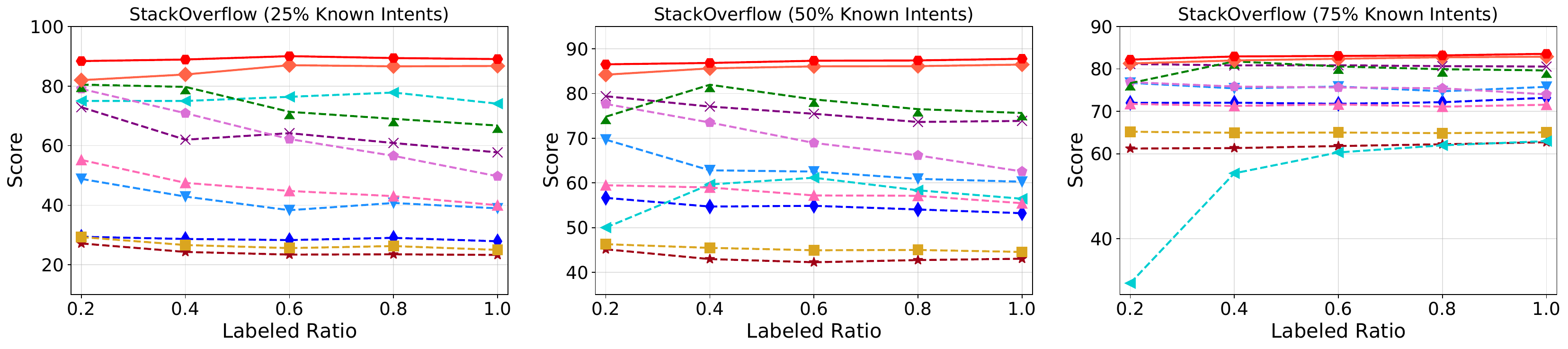}
	\includegraphics[width=1.7\columnwidth ]{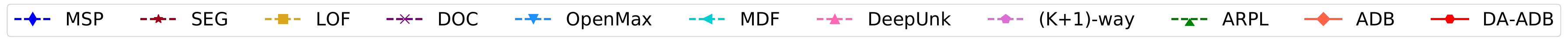}
	\caption{Influence of the labeled ratio on the three datasets with different known class ratios.}
	\label{Aba_3}
\end{figure*}

\subsection{Effect of Distance-aware Representation Learning}
\subsubsection{Visualization of Intent Representations}
In Fig~\ref{feats_visualization}, we use t-SNE~\cite{van2008visualizing} to visualize the learned intent representations on the testing set of the BANKING dataset. It is shown that the vanilla intent representations of known intents are dispersed, which may result in fuzzy boundaries between adjacent clusters. The open intent representations are distributed haphazardly throughout the feature space, which easily leads to confusion between open and known intent samples. 

In contrast, the intent representations learned with distance-aware concepts are more discriminative. The known intent samples are with intra-class compactness and inter-class separation properties. Surprisingly, the distributions of open intent samples are more concentrated to be far away from many known intents, which is beneficial to be better detected.  
\subsubsection{Analysis of Hyper-parameter $\alpha$}
\label{analysis_of_alpha}
In Fig~\ref{Aba_4}, we use the accuracy score as the metric and show the effect of $\alpha$ on three datasets with different known class ratios. The hyper-parameter $\alpha$ is used to control the logits of the cosine classifier within a desirable range. Intuitively, a large $\alpha$ is better as it can scale the cosine similarity scores and further increase the peakiness of the softmax distribution~\cite{gidaris2018dynamic} to enhance the discrimination of the intent representations. 

However, it is interesting to observe that though a larger $\alpha$ may achieve higher performance with fewer known intents ($25\%$), the performance drops rapidly with more known intents (75$\%$). We suppose the reason is that the discriminative feature distributions help learn compact decision boundaries, which works better when there are fewer known intent samples. Nevertheless, it may not be good to identify more known intent samples with tight decision boundaries.

\subsection{Effect of Adaptive Decision Boundary Learning}
\subsubsection{Analysis of Radius of Decision Boundary $\Delta$}
To verify the discrimination of the learned decision boundary, we use different ratios of $\Delta$ (radius of the decision boundary) for open intent detection on the BANKING dataset and show the results in Fig~\ref{boundary}. The dotted lines indicate the performance with our learned $\Delta$. 

DA-ADB achieves the best or competitive performance with $\Delta$ among all assigned decision boundaries, which verifies the tightness of the learned decision boundary. Though $1.25\Delta$ is slightly better on 75$\%$ known classes, it is lower on the other two settings. We notice that the performance of open intent detection is sensitive to the size of the decision boundaries. Overcompact decision boundaries will increase the open space risk by misclassifying more known intent samples as the open intent. Correspondingly, overrelaxed decision boundaries will increase the empirical risk by misclassifying more open intent samples as known intents. The two cases both perform worse compared with $\Delta$.
\subsubsection{Boundary Learning Process}
Fig~\ref{Aba_2_1} shows the decision boundary learning process. At first, most parameters are assigned small values near zero after initialization, which leads to a small radius with the $\operatorname{Softplus} $ activation function. As the initial radius is too small, the empirical risk is dominant. Therefore, the radius of each decision boundary expands to contain more known intent samples belonging to its class. As the training process goes on, the radius of the decision boundary learns to be large enough to contain most of the known intents. However, the large radius will also introduce redundant open intent samples. In this case, the open space risk plays a dominant role, which prevents the radius from enlarging. Finally, the decision boundaries converge with a balance between empirical and open space risks.
\subsection{Effect of Labeled Data}
To investigate the influence of labeled data, we vary the labeled ratio in training set to 0.2, 0.4, 0.6, 0.8, and 1.0. The accuracy score is used to evaluate the performance. As shown in Fig~\ref{Aba_3}, DA-ADB outperforms all the other baselines on three datasets in almost all settings. Besides, it keeps a more robust performance under different labeled ratios than other methods. 

Notably, the probability-based methods (e.g., MSP, DOC, OpenMax, and ($K$+1)-way) show better performance with less labeled data in many cases. We suppose the reason is that the predicted scores are in low confidence with less prior knowledge for training, which is helpful to reject the open intent with the threshold. However, as the number of labeled data increases, these methods tend to be biased towards the known intents with the aid of the strong feature extraction capability of DNNs~\cite{7298640} and suffer performance degradation. We also notice that the performance of OpenMax and ARPL is unstable. The former computes centroids of each known class with only corrective positive training samples, and the labeled ratio may easily influence the qualities of centroids. The latter faces a larger open space risk when learning reciprocal points with less prior knowledge of labeled data. Compared with the methods mentioned above,  ($K$+1)-way achieves more robust performance by using pseudo samples as the open class.

In addition, the feature-based methods (e.g., SEG, LOF, DeepUnk)  adopt a density-based novelty detection algorithm to perform open intent detection. These methods largely depend on the prior knowledge of labeled data, and their performance all drop dramatically with less labeled data. MDF employs an SVM-based method to detect outliers using Mahalanobis distance information, but it is also sensitive to the amount of labeled data and has limitations in leveraging the prior knowledge, particularly with more known intents. ADB shows excellent and robust performance with appropriate learned decision boundaries, but it performs worse in many settings without utilizing the distance information during feature learning. 

\section{Conclusions}
This article focuses on a substantial problem, open intent detection in NLU. This problem uses only known intents as prior knowledge, and the goal is not only to identify these known intents but also to detect the open intent. Obtaining friendly intent representations and appropriate decision boundaries are two critical challenges for open intent detection. To solve these problems, we propose a novel pipeline framework, DA-ADB, which learns discriminative features with distance-aware concepts and learns suitable decision boundaries by balancing both empirical and open space risks. We conducted extensive experiments on three benchmark datasets to show the superiority of the proposed method. Our approach yields significant improvements over state-of-the-art methods and achieves robust performance with different known intents and labeled data ratios.

\section*{Acknowledgments}
Sincerely, we thank the help and constructive feedback of Ting-En Lin.

\bibliographystyle{IEEEtran}
\bibliography{TASLP}.

 \begin{IEEEbiography} 
  [{\includegraphics[width=1in,height=1.25in,clip,keepaspectratio]{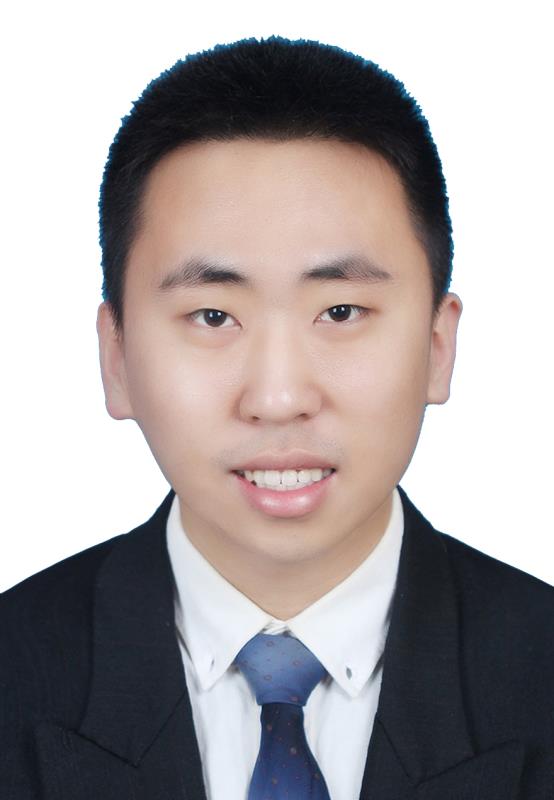}}]{Hanlei Zhang}
received the B.S. degree from the
Department of Computer Science and Technology,
Beijing Jiaotong University, Beijing, China, in 2020.
He is currently working toward the Ph.D. degree
with the Department of Computer Science and Technology, Tsinghua University, Beijing, China. He has authored or coauthored five peer-reviewed papers in top-tier international conferences, including AAAI, ACM MM, and ACL. His research interests include intent analysis, open world classification, clustering, multimodal language understanding, natural
language processing.
 \end{IEEEbiography}

 \begin{IEEEbiography}    [{\includegraphics[width=1in,height=1.25in,clip,keepaspectratio]{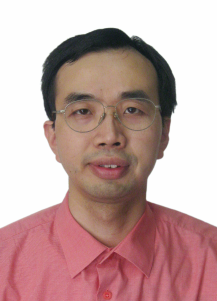}}]{Hua Xu} received the B.S. degree from Xi’an Jiaotong University, Xi’an, China, in 1998, and the M.S. and Ph.D. degrees from Tsinghua University, Beijing, China, in 2000 and 2003, respectively. He is a Tenured Associate Professor with the Department of Computer Science and Technology, Tsinghua University. He has authored or coauthored more than 130 peer-reviewed papers in top-tier international journals and conferences. His research interests include multi-modal intelligent information processing for natural interaction of service robots, evolutionary learning, and intelligent optimization. Prof. Xu was the recipient of the Second Prize from the National Science and Technology Progress of China, First Prize from Beijing Science and Technology, and Third Prize from Chongqing Science and Technology.
 \end{IEEEbiography}

 \begin{IEEEbiography} 
  [{\includegraphics[width=1in,height=1.15in,clip,keepaspectratio]{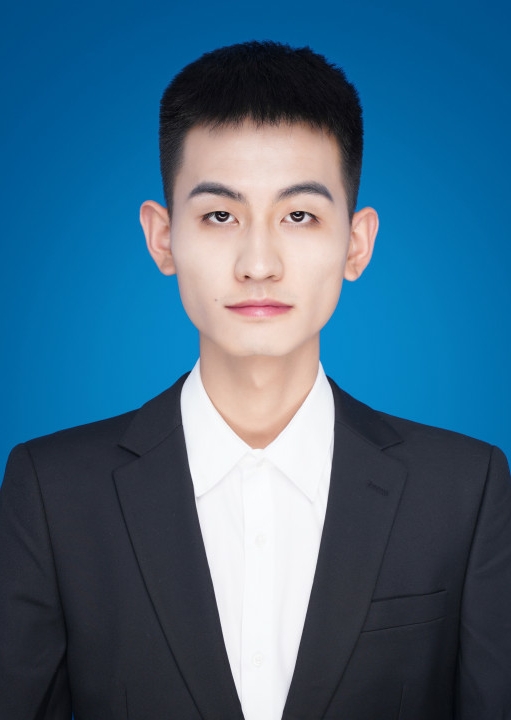}}]{Shaojie Zhao}
is currently working toward the M.S. degree with the School of Information Science and
Engineering, Hebei University of Science and Technology, Shijiazhuang, China. He has authored or coauthored one paper in the ACM MM international conference. His research interests include intent detection, open world classification, and natural language processing.
 \end{IEEEbiography}

 \begin{IEEEbiography} 
  [{\includegraphics[width=1in,height=1.25in,clip,keepaspectratio]{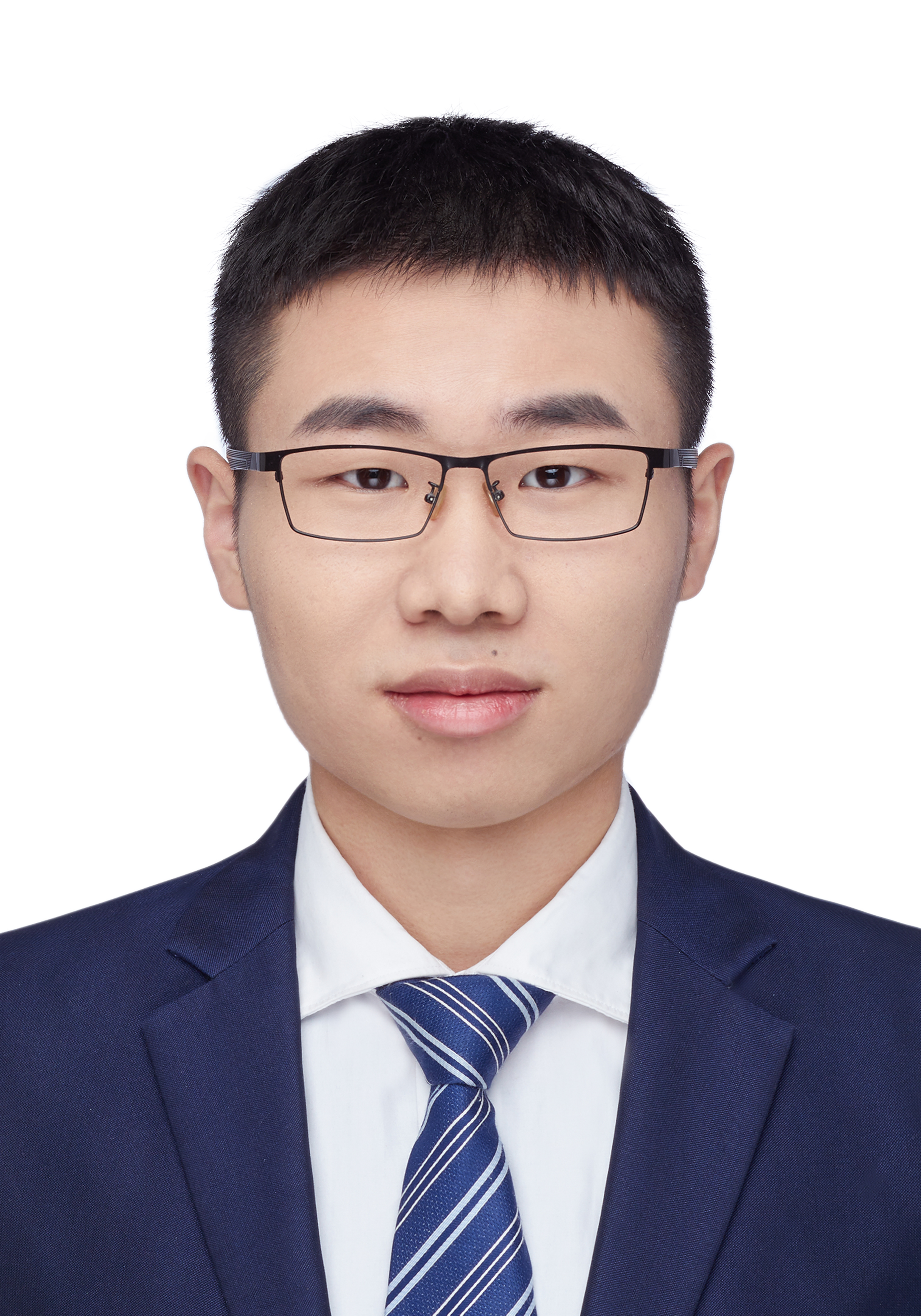}}]{Qianrui Zhou} received the B.S. degree in 2022 from
the Department of Computer Science and Technology, Tsinghua University, Beijing, China, where he is currently working toward the Ph.D. degree with the Department of Computer Science and Technology. He has authored or coauthored two papers in international conferences, including ACM MM and ACL. His research interests include natural language
processing and multimodal machine learning.
 \end{IEEEbiography}
 





\end{document}